\documentclass[letterpaper, 10 pt, conference]{ieeeconf}  

\IEEEoverridecommandlockouts                              

\usepackage{graphicx}
\usepackage{subfigure}
\usepackage{amsmath}
\usepackage{amsfonts,amssymb}
\usepackage{cite}
\usepackage{bm}
\usepackage{cases}
\usepackage[utf8]{inputenc}
\usepackage{makecell}
\usepackage{hyperref}

\usepackage{soul}
\usepackage{color, xcolor} 

\title{\LARGE \bf
Calibration of Deep Learning Classification Models in fNIRS
}

\author{Zhihao Cao$^{1, \dag, \ddag}$, Zizhou Luo$^{2, \dag}$
\thanks{$^{1}$Zhihao Cao is with the Department of Mathematics, 
        ETH Zurich, Switzerland. Email: {\tt\small zhicao@ethz.ch}}%
\thanks{$^{2}$Zizhou Luo is with the Department of Informatics, 
        University of Zurich, Switzerland. Email: {\tt\small zizhou.luo@uzh.ch}}%
\thanks{$^{\dag}$Authors have contributed equally to the article.}%
\thanks{$^{\ddag}$Corresponding author. Email: {\tt\small zhicao@ethz.ch}}%
\thanks{$^{\S}$\href{https://github.com/WhiteFireFox/fNIRS-Calibration}{https://github.com/WhiteFireFox/fNIRS-Calibration}}%
}
\begin{document}

\maketitle
\thispagestyle{empty}
\pagestyle{empty}

\begin{abstract}
Functional near-infrared spectroscopy (fNIRS) is a valuable non-invasive tool for monitoring brain activity. The classification of fNIRS data in relation to conscious activity holds significance for advancing our understanding of the brain and facilitating the development of brain-computer interfaces (BCI). Many researchers have turned to deep learning to tackle the classification challenges inherent in fNIRS data due to its strong generalization and robustness. In the application of fNIRS, reliability is really important, and one mathematical formulation of the reliability of confidence is calibration. However, many researchers overlook the important issue of calibration. To address this gap, we propose integrating calibration into fNIRS field and assess the reliability of existing models. Surprisingly, our results indicate poor calibration performance in many proposed models. To advance calibration development in the fNIRS field, we summarize three practical tips. Through this letter, we hope to emphasize the critical role of calibration in fNIRS research and argue for enhancing the reliability of deep learning-based predictions in fNIRS classification tasks. All data from our experimental process are openly available on GitHub$^{\S}$.
\end{abstract}

\section{INTRODUCTION \label{sec:introduction}}
There are various methods for humans to understand brain signals, including electroencephalogram (EEG) \cite{buzsaki2012origin}, functional magnetic resonance imaging (fMRI) \cite{logothetis2008we}, and functional near-infrared spectroscopy (fNIRS) \cite{jobsis1977noninvasive}. Functional Near-Infrared Spectroscopy (fNIRS) has become valuable for non-invasive monitoring of brain activity owing to its portability and reduced susceptibility to electrical noise and motion artifacts \cite{naseer2015fnirs}. By detecting changes in oxygenated hemoglobin (HbO) and deoxygenated hemoglobin (HbR) concentrations via near-infrared light absorption, fNIRS facilitates the examination of human brain activity \cite{jobsis1977noninvasive}. Understanding of fNIRS signals, particularly the classification of human behavioral intentions associated with these signals, holds promise for the advancement of brain-computer interfaces (BCIs) and providing assistance for individuals with disabilities \cite{ferrari2012brief}.

Currently, various effective methods exist for classifying fNIRS signals, including traditional machine learning and deep learning. Shin et al. \cite{shin2016open} introduced Linear Discriminant Analysis (LDA), mapping fNIRS signals to distinct brain functional states. Chen et al. \cite{chen2020classification} proposed a feature extraction method based on the general linear model (GLM) to identify features from sparse fNIRS channels. However, traditional machine learning methods heavily rely on manual feature selection and prior knowledge, often resulting in poorer generalization compared with deep learning. Consequently, there is a growing trend in leveraging robust deep learning methods for fNIRS classification. Lyu et al. \cite{lyu2021domain} applied CNN-based and LSTM-based methods to extract fNIRS signal characteristics. Furthermore, a 1D-CNN neural network \cite{sun2020novel} structure was employed to extract features from NIRS signals, enhancing classification accuracy by Sun et al. Rojas et al. \cite{rojas2021pain} utilized a Bi-LSTM model to automatically extract features from raw fNIRS data. What's more, Wang et al. \cite{wang2022transformer} proposed a transformer-based classification network (fNIRS-T), to enhances network representation capabilities by exploring spatial-level and channel-level signal representations. Cao et al. \cite{cao2024simple} proposed integrating metric learning into fNIRS research to improve networks capability in excluding out-of-distribution outliers. Additionally, Wang et al. \cite{wang2023rethinking} proposed incorporating delayed hemodynamic response as domain knowledge into fNIRS classification to enhance classification accuracy.

\begin{figure}[htpb]
  \centering
  \vspace{-0.3cm}
  \subfigure[]{
  \includegraphics[width=1.55in]{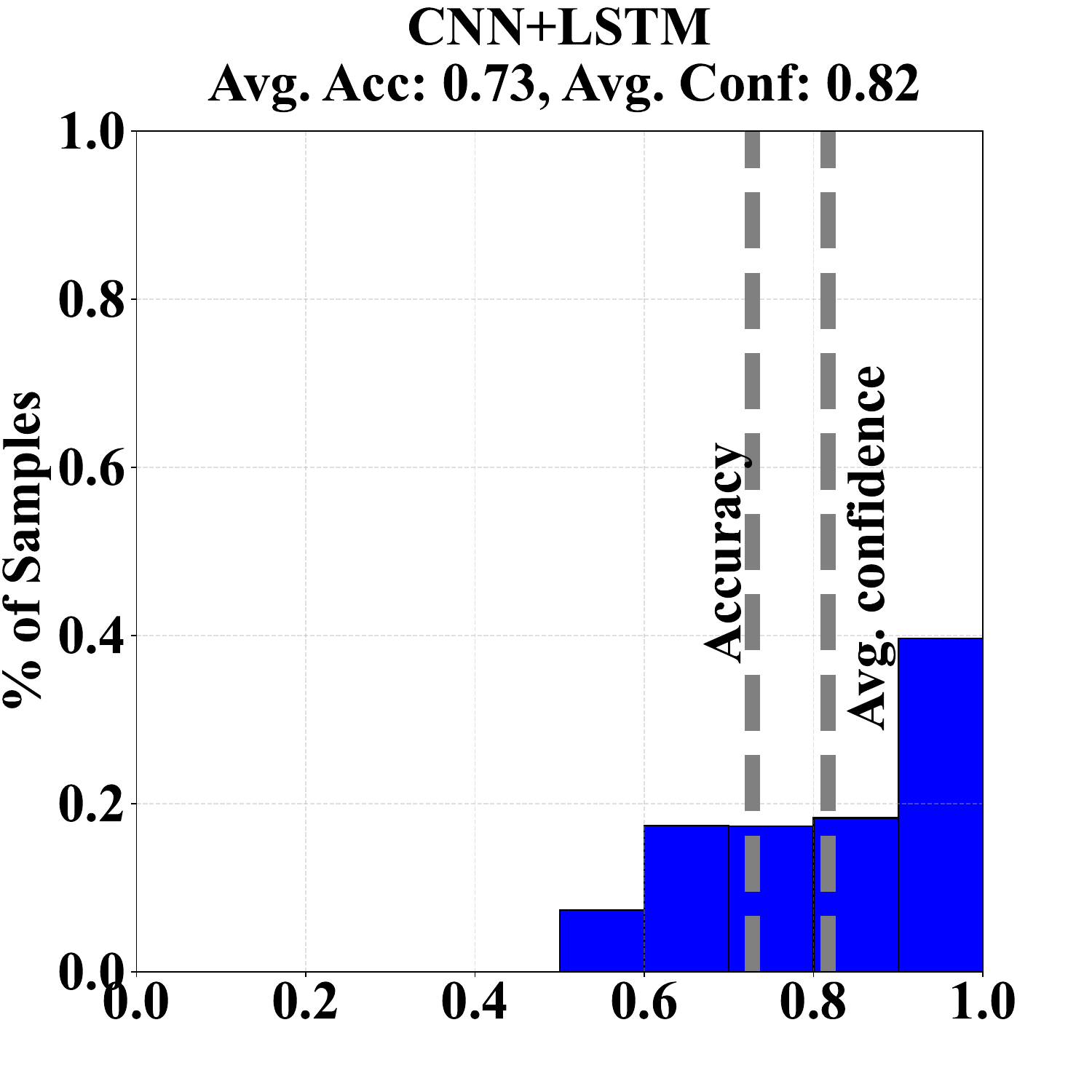}
  }
  \subfigure[]{
  \includegraphics[width=1.55in]{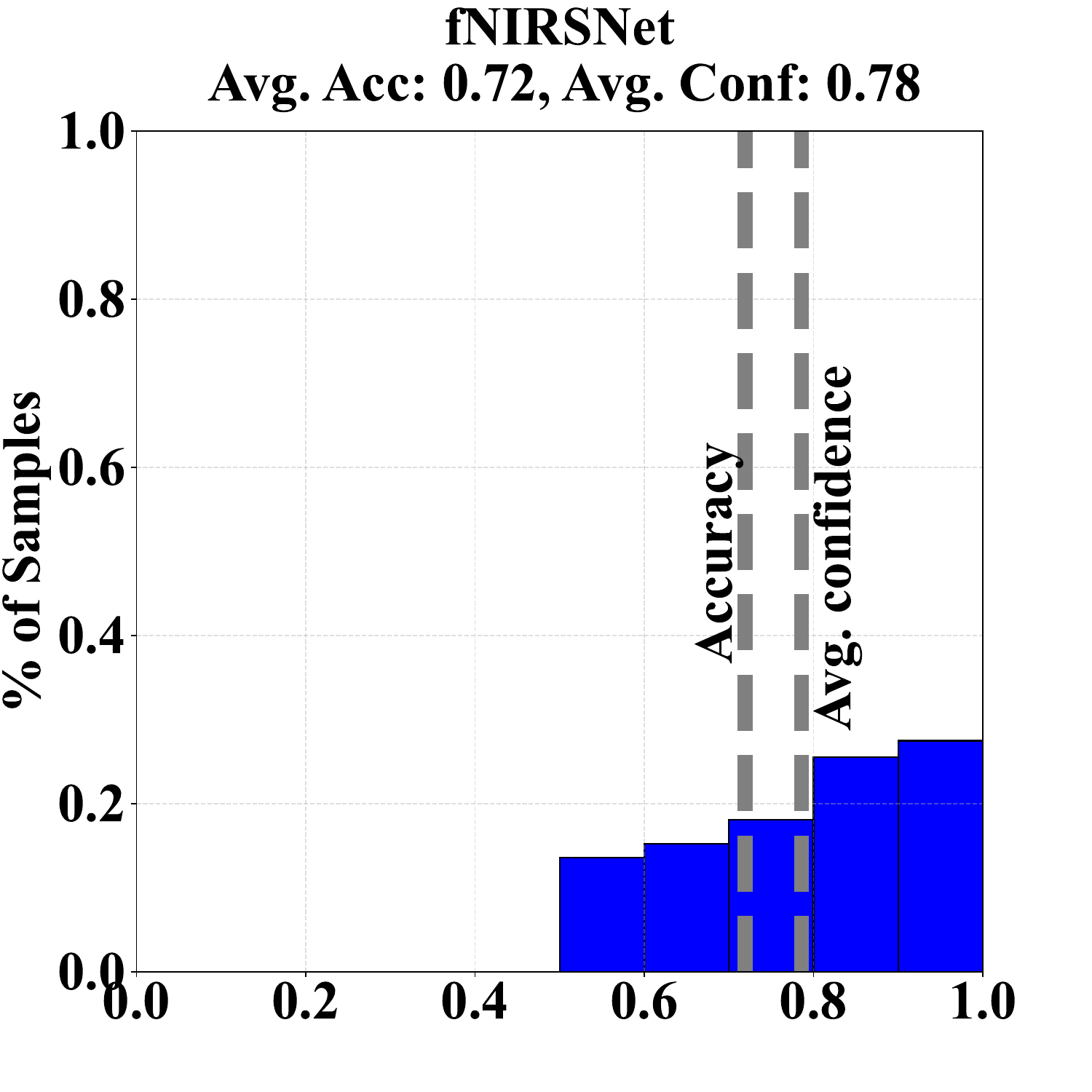}
  }
  \subfigure[]{
  \includegraphics[width=1.55in]{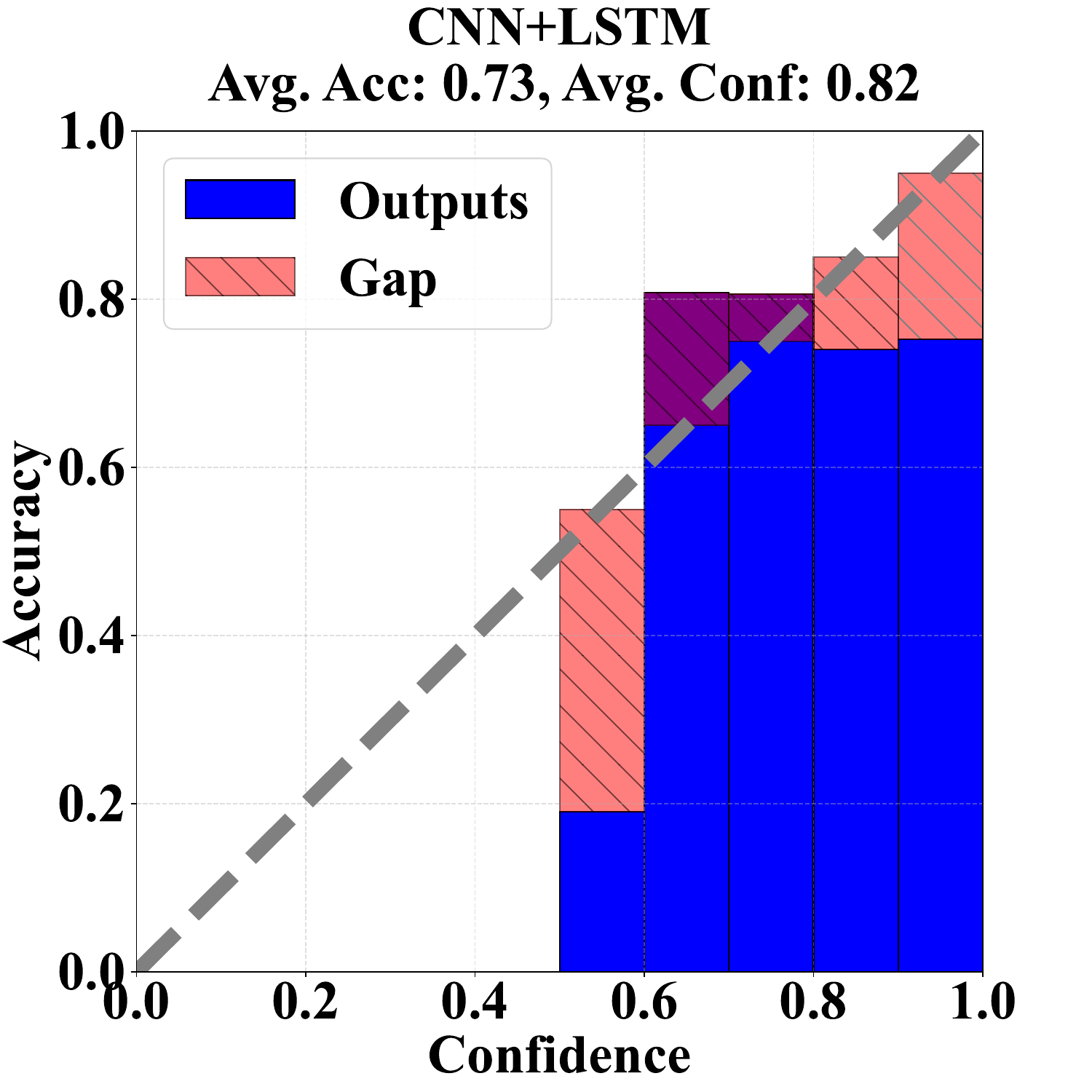}
  }
  \subfigure[]{
  \includegraphics[width=1.55in]{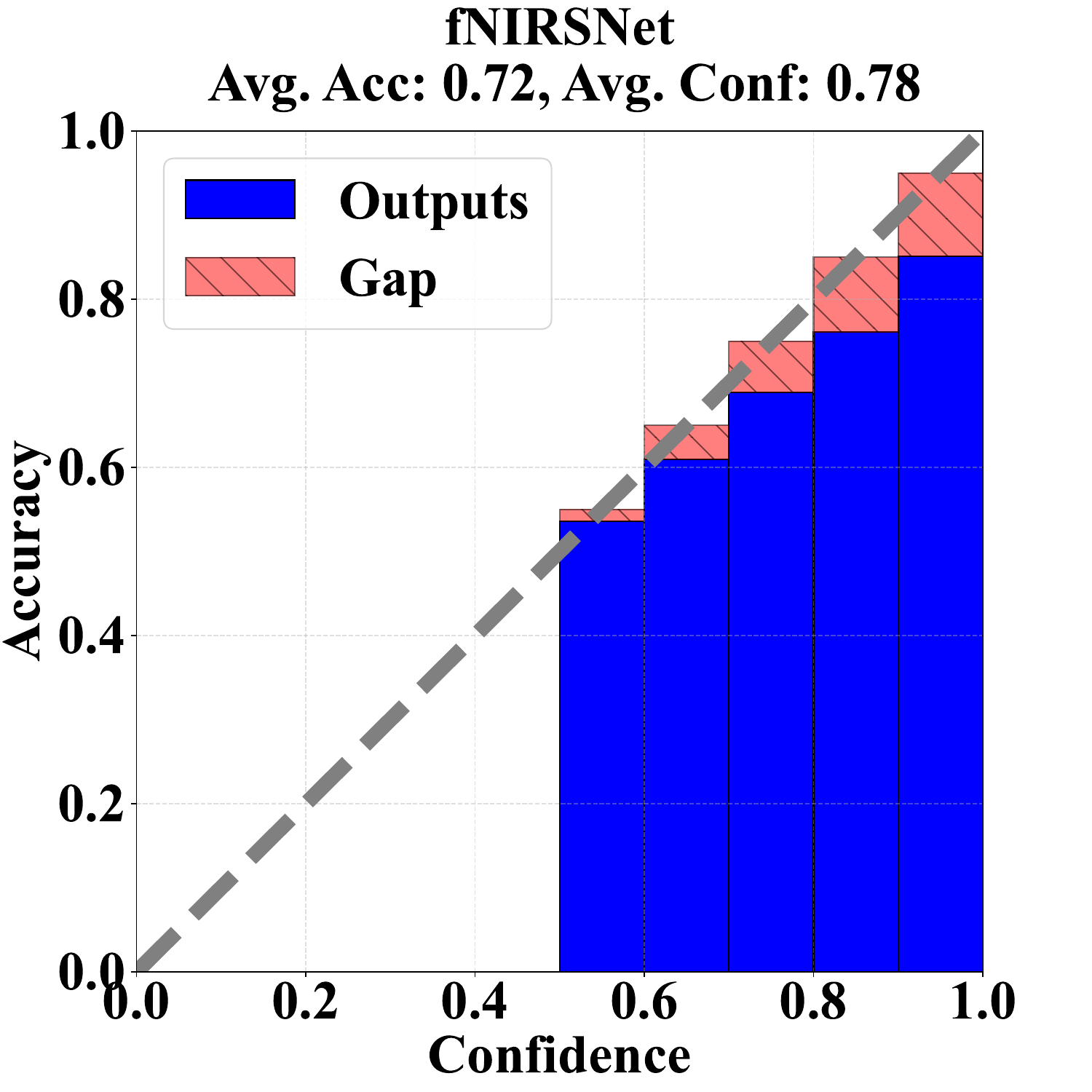}
  }
  \vspace{-0.1cm}
  \caption{Examining the results from the last epoch of each cross-validation, and observing nearly similar accuracy and confidence levels between (a) \cite{mughal2021fnirs} and (b) \cite{wang2023rethinking}. However, the calibration of (b) surpasses that of (a). Section \ref{sec:experiment} further validates these findings, emphasizing the important role of calibration}
  \label{fig:intro_calibration}
  \vspace{-0.1cm}
\end{figure}

Reliability is crucial in the application of fNIRS. Classification networks in fNIRS should not only prioritize accuracy but also indicate potential inaccuracies \cite{guo2017calibration}. An example illustrating high accuracy but low reliability of fNIRS classification networks can be seen in Figure \ref{fig:intro_calibration}. One mathematical formulation of the reliability of confidence is calibration \cite{dawid1982well}, which is important for classification models in various fields, ensuring probability estimates align closely with correctness likelihood. Guo et al. \cite{guo2017calibration} applied Reliability Diagrams and Expected Calibration Error (ECE) to express calibration issues. Despite the effectiveness of ECE, challenges still remain \cite{nixon2019measuring}, including fixed calibration ranges, trade-off between two variances and so on. To address these challenges, Nixon et al. \cite{nixon2019measuring} introduced novel calibration metrics: Static Calibration Error (SCE), Adaptive Calibration Error (ACE) and Thresholded Adaptive Calibration Error (TACE). What's more, there still exists Maximum Calibration Error (MCE) \cite{naeini2015obtaining} and Overconfidence Error (OE) \cite{thulasidasan2019mixup} to measure calibration, which we will discuss in the section \ref{sec:metrics} and section \ref{sec:experiment}. 

Through assessing the calibration error of the models, we discover that some existing network models used for analysing fNIRS exhibit poor calibration, indicating low reliability. That is to say, many of these approaches overlook the confidence calibration problem \cite{guo2017calibration}, which is essential for ensuring the reliability of predictions and directly reflects the likelihood of its ground truth correctness. Hence, we hope to emphasize the critical role of calibration in fNIRS research and argue for enhancing the reliability of deep learning-based predictions in fNIRS classification tasks. We also summarize three practical tips, including the balance of accuracy and calibration, the select of network and temperature scaling for calibration, in order to advance calibration development in the fNIRS field. The contributions of this letter are summarized as follows.

\begin{itemize}

\item Establishing a benchmark for evaluating the reliability of deep learning in fNIRS classification tasks and extensively verifying the calibration errors of some current deep learning models in fNIRS classification tasks through our experiments.

\item Summarizing and experimentally verifying three practical tips, including balancing accuracy and calibration, appropriate networks selection, and temperature scaling for calibration.

\end{itemize}

The rest of this letter is organized as follows. Section \ref{sec:dataset} presents the dataset utilized in this study and notations of the fNIRS dataset. In Section \ref{sec:metrics}, we discuss various metrics for evaluating model calibration performances along with their formal expressions. Section \ref{sec:experiment} describes the dataset's preprocessing methods, the experimental setup and calibration evaluation result. Section \ref{sec:skills} summarizes three practical tips, including the balance of accuracy and calibration, the select of network and temperature scaling for calibration. Finally, Section \ref{sec:conclusion} provides a summary and conclusion of this letter.

\section{FUNCTIONAL NEAR-INFRARED SPECTROSCOPY DATASET \label{sec:dataset}}

To ensure reproducibility and diversity, our experiments utilize two open-source datasets. These datasets encompass the mental arithmetic classification task \cite{shin2016open} and the unilateral finger and foot tapping classification task \cite{bak2019open}.

\subsection{Mental Arithmetic (MA)} 

The Mental Arithmetic (MA) \cite{shin2016open} dataset was collected using NIRScout at a sampling rate of 12.5 Hz. Fig. \ref{fig:MA} (a) illustrates the placement of fourteen sources and sixteen detectors. Each trial comprised an introductory period (2 s), a task period (10 s) and an inter-trial rest (15–17 s) and concluding with a short beep (250 ms), as depicted in Fig. \ref{fig:MA} (b). During the task, subjects were instructed to perform two actions (mental arithmetic and baseline) randomly displayed on the screen.

\begin{figure}[htpb]
  \centering
  \subfigure[]{
  \includegraphics[width=1.55in]{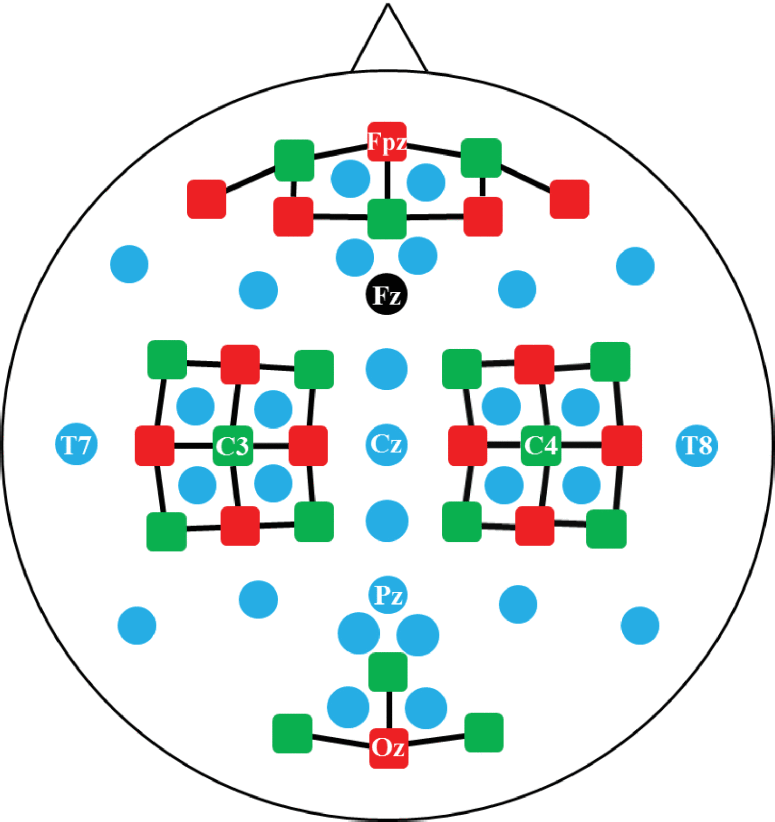}
  }
  \subfigure[]{
  \includegraphics[width=1.55in]{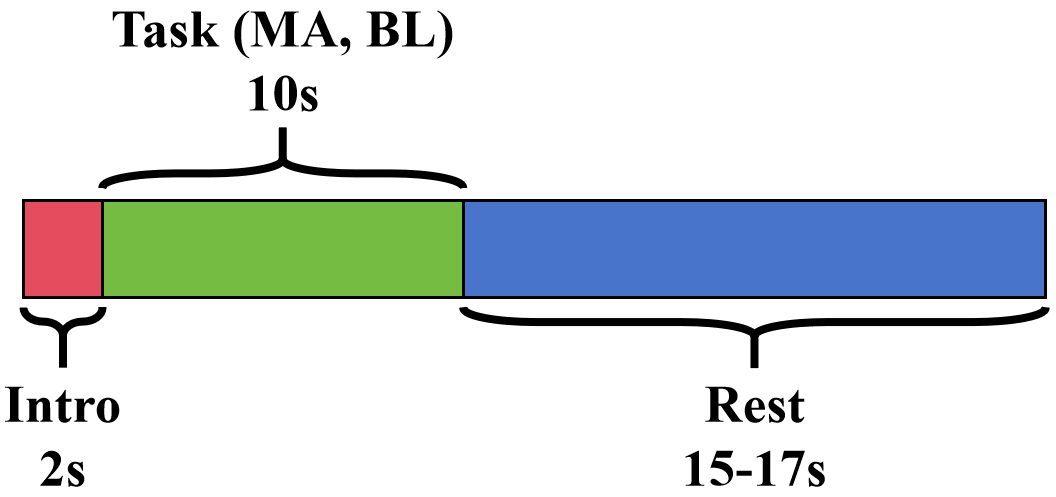}
  }
  \vspace{-0.1cm}
  \caption{(a) Sensor location layout for MA \cite{shin2016open}. (b) MA's trial consists of an introduction period, a task period, and a rest period.}
  \label{fig:MA}
  \vspace{-0.5cm}
\end{figure}

\subsection{Unilateral Finger and Foot Tapping (UFFT)} 

The Unilateral Finger and Foot Tapping (UFFT) \cite{bak2019open} dataset was recorded using a three-wavelength continuous-time multi-channel fNIRS system, comprising Tx and Rx, as depicted in Fig. \ref{fig:UFFT} (a). Each trial consisted of an introduction period (2 s), a task period (10 s), and an inter-trial rest (17–19 s), as depicted in Fig. \ref{fig:UFFT} (b). During the task, subjects were instructed to perform three distinct actions (RHT, LHT and FT) based on on-screen instructions.

\begin{figure}[htpb]
  \centering
  \subfigure[]{
  \includegraphics[width=1.55in]{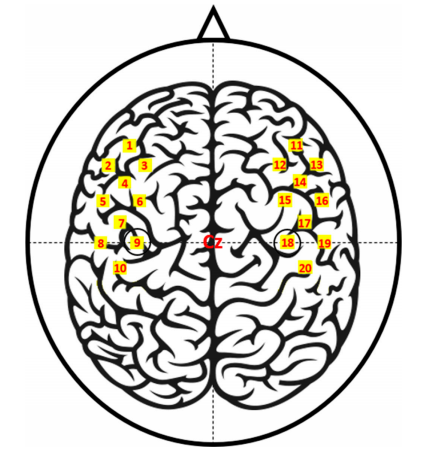}
  }
  \subfigure[]{
  \includegraphics[width=1.55in]{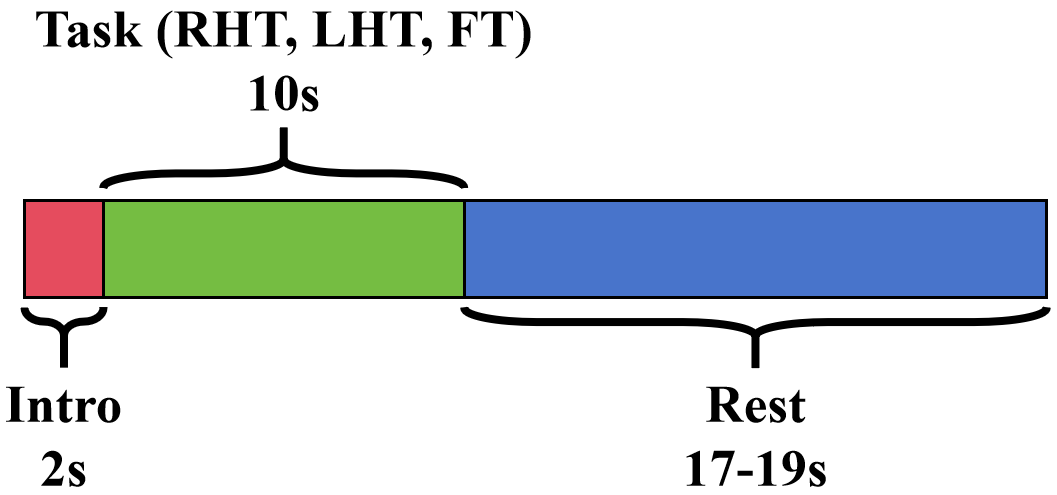}
  }
  \vspace{-0.1cm}
  \caption{(a) Sensor location layout for UFFT \cite{bak2019open}. (b) UFFT's trial consists of an introduction period, a task period, and a rest period.}
  \label{fig:UFFT}
  \vspace{-0.5cm}
\end{figure}

\section{CALIBRATION ERROR \label{sec:metrics}}


Calibration error is a measure of the degree of deviation between confidence and accuracy, indirectly reflecting the trustworthiness of the classification result. several scalar statistics have been proposed \cite{guo2017calibration,nixon2019measuring,naeini2015obtaining,thulasidasan2019mixup}.

\subsection{Expected Calibration Error (ECE)} 

ECE approximates calibration error in expectations by discretizing the probability interval into bins and assigning each predicted probability to its respective bin. The calibration error is calculated as the disparity between the accuracy (fraction of correct predictions) and the mean of the probabilities within each bin.

\begin{equation}
\label{ece_equation}
ECE = \sum_{b=1}^{B}
\frac{n_{b}}{N}|acc(b)-conf(b)|\;,
\end{equation}
where $n_b$ is the number of predictions in bin $b$, $N$ is the total number of data points, and $acc(b)$ and $conf(b)$ are the accuracy and confidence of bin $b$, respectively.

\subsection{Maximum Calibration Error (MCE)} 

MCE minimizes the maximum deviation between confidence and accuracy, particularly important for high-risk applications. It computes the distinction between predicted probabilities and true labels for all bins, identifying the maximum value among them.

\begin{equation}
\label{ece_equation}
MCE = \max(|acc(b)-conf(b)|)\;,
\end{equation}
where $\max(\cdot)$ is the maximum value in the set, and $acc(b)$ and $conf(b)$ are the accuracy and confidence of bin $b$, respectively.

\subsection{Overconfidence Error (OE)} 

OE deals with high-risk scenarios where confident yet incorrect predictions can be particularly harmful. It penalizes predictions using confidence weights when confidence exceeds accuracy, imposing a significant penalty on overconfident bins.

\begin{equation}
\label{ece_equation}
OE = \sum_{b=1}^{B}
\frac{n_{b}}{N}(conf(b)\times \max(conf(b) - acc(b), 0))\;,
\end{equation}
where $\max(\cdot)$ is the maximum value in the set, $n_b$ is the number of predictions in bin $b$, $N$ is the total number of data points, and $acc(b)$ and $conf(b)$ are the accuracy and confidence of bin $b$, respectively.

\subsection{Static Calibration Error (SCE)} 

SCE is introduced as a straightforward extension of expected calibration error for addressing multi-class problems. SCE bins predict probabilities for each class individually, compute calibration error within each bin, and then average across bins.

\begin{equation}
\label{sce_equation}
SCE = \frac{1}{K}\sum_{k=1}^{K}\sum_{b=1}^{B}
\frac{n_{bk}}{N}|acc(b, k)-conf(b, k)|\;,
\end{equation}
where $acc(b, k)$ and $conf(b, k)$ are the accuracy and confidence of bin $b$ for class label $k$, respectively, $n_{bk}$ is the number of predictions in bin $b$ for class label $k$, and $N$ is the total number of data points.

\subsection{Adaptive Calibration Error (ACE)} 

ACE is introduced to achieve the most accurate estimation of the overall calibration error. ACE adapts calibration ranges based on the bias-variance trade-off, emphasizing regions with frequent predictions while downplaying less probable regions. It divides bin intervals to ensure an equal number of predictions in each interval.

\begin{equation}
\label{ace_equation}
ACE = \frac{1}{KR}\sum_{k=1}^{K}\sum_{r=1}^{R}
\frac{n_{bk}}{N}|acc(r, k)-conf(r, k)|\;,
\end{equation}
where $acc(r, k)$ and $conf(r, k)$ are the accuracy and confidence of adaptive calibration range $r$ for class label $k$, respectively, $N$ is the total number of data points, and calibration range $r$ defined by the $N/R_{th}$ index of the sorted and thresholded predictions.

\subsection{Adaptive Calibration Error (TACE)} 

In cases where there are many classes, Softmax predictions can result in infinitesimal probabilities, which may wash out the calibration score. Thus, TACE computes values above the threshold, and the calculation process is similar to ACE, with the threshold set at 0.01.

\section{EXPERIMENT \label{sec:experiment}}

\subsection{Signal Preprocessing} 

In the MA dataset, signal preprocessing involved calculating changes in deoxygenated hemoglobin and oxyhemoglobin (HbR and HbO) concentrations with the modified Beer-Lambert law \cite{cope1988system}. Subsequently, HbO and HbR data were filtered using a sixth-order zero-phase Butterworth band-pass filter with a passband of 0.01–0.1 Hz. The data are segmented using a sliding window (window size = 3 s, step size = 1 s) \cite{shin2016open}, and subsequently normalized.

In the UFFT dataset, HbO and HbR data were filtered using a third-order zero-phase Butterworth band-pass filter with a passband of 0.01–0.1 Hz. Baseline correction is performed by subtracting the average value from the period between -1 and 0 seconds. The data are also segmented using a sliding window (window size = 3 s, step size = 1 s) \cite{sun2020novel}, and then normalized.

\subsection{Training Setting} 

We conducted experiments on a workstation equipped with a 12th Gen Intel(R) Core(TM) i9-12900H processor and an NVIDIA GeForce RTX 3080 Ti GPU with 32GB of RAM. The settings for the optimizer, loss function, number of training epochs, learning rate, and other parameters in the training process were designed following the methodology outlined by Wang et al. \cite{wang2023rethinking}.

1D-CNN \cite{sun2020novel}, CNN \cite{lyu2021domain}, LSTM \cite{lyu2021domain}, CNN+LSTM \cite{mughal2021fnirs}, and fNIRSNet \cite{wang2023rethinking} maintain the original network structure with minimal alterations, primarily adjusting input and output dimensions for data accommodation and classification purposes. For fNIRS-T \cite{wang2022transformer}, following the specifications in \cite{wang2022transformer, wang2023rethinking}, $Conv_S$ and $Conv_C$ kernel sizes are set to 5 × 10 and 1 × 10, respectively. The Transformer layers of fNIRS-T are configured as 4, with the linear projection and MLP layer dimension set to 32.

\subsection{Evaluation Setting}

Deep learning model are trained for each subject using a 5-fold cross-validation (KFold-CV) that splits training and test sets according to trials to avoid information leakage \cite{wang2023rethinking}.

\subsection{Calibration Result} 

According to Table \ref{tab:ma_result}, on the MA dataset, both deep learning networks, fNIRSNet and CNN+LSTM, show high accuracy. However, CNN+LSTM notably demonstrates lower calibration compared to fNIRSNet. Despite LSTM's lower accuracy, it performs well in calibration error indicators: SCE, ACE, and TACE. The prediction results of LSTM and fNIRSNet on the MA dataset are visualized in Figure \ref{fig:MA}.

It's clear that LSTM's calibration (SCE, ACE, and TACE) surpasses fNIRSNet due to more LSTM predictions falling into low-confidence bins. This results in improved calibration performance for LSTM within low-confidence bins, thus exceeding fNIRSNet in terms of SCE, ACE, and TACE.

\begin{figure}[htpb]
  \centering
  \vspace{-0.5cm}
  \subfigure[LSTM on MA Dataset]{
  \includegraphics[width=1.55in]{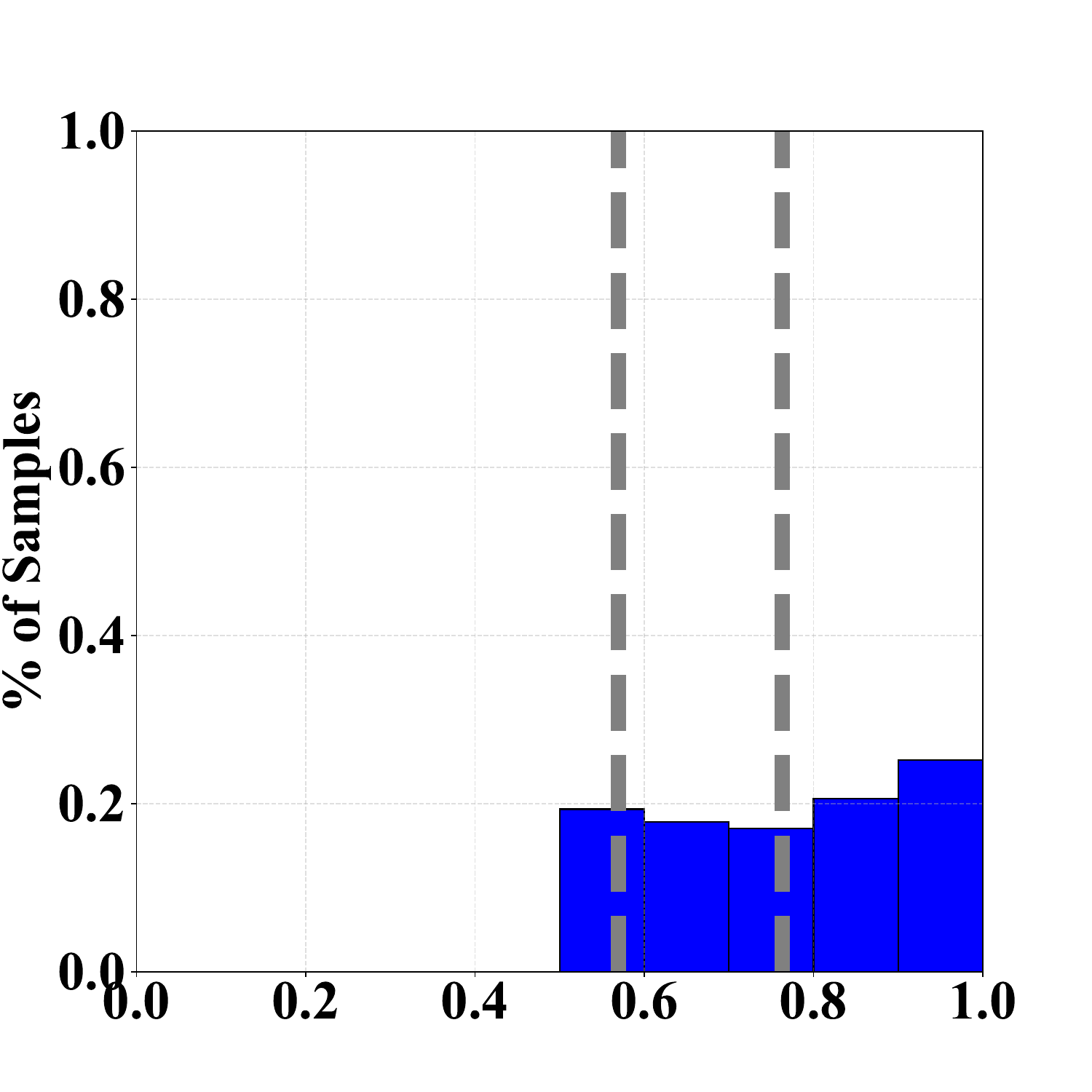}
  }
  \subfigure[fNIRSNet on MA Dataset]{
  \includegraphics[width=1.55in]{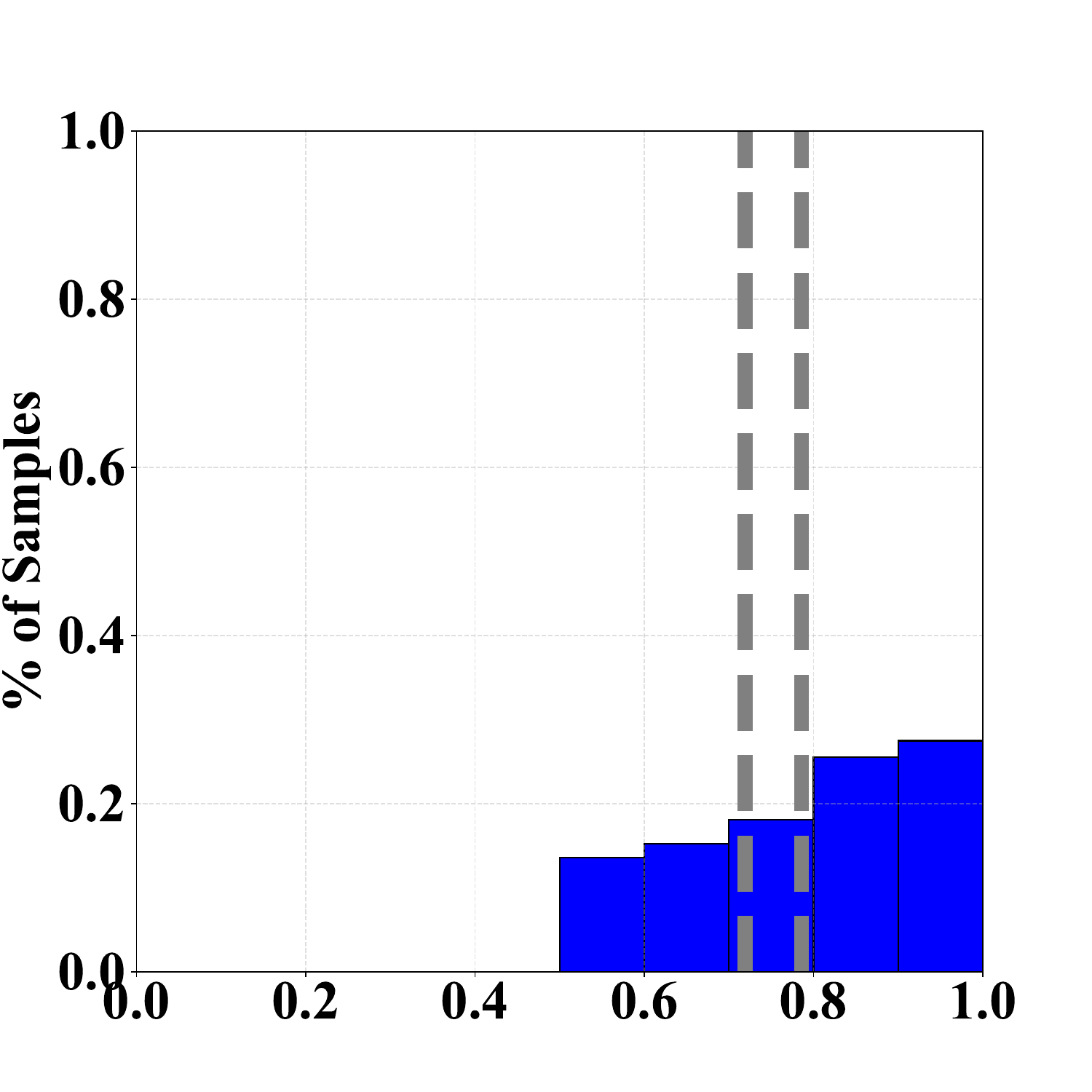}
  }
  \vspace{-0.1cm}
  \caption{LSTM and fNIRSNet on MA Dataset}
  \label{fig:MA}
  \vspace{-0.3cm}
\end{figure}

According to Table \ref{tab:ufft_result}, on the UFFT dataset, fNIRSNet shows high accuracy and well-calibration (OE, SCE, ACE, and TACE), while CNN+LSTM performs better on ECE and MCE indicators. We visualize the prediction results of CNN+LSTM and fNIRSNet on the UFFT dataset in Figure \ref{fig:UFFT}. Considering Table \ref{tab:ufft_result}, it's evident that fNIRSNet's calibration performance (SCE, ACE, and TACE) surpasses CNN+LSTM due to a larger proportion of fNIRSNet predictions falling into low-confidence bins.

\begin{figure}[htpb]
  \centering
  \vspace{-0.5cm}
  \subfigure[CNN+LSTM on UFFT Dataset]{
  \includegraphics[width=1.55in]{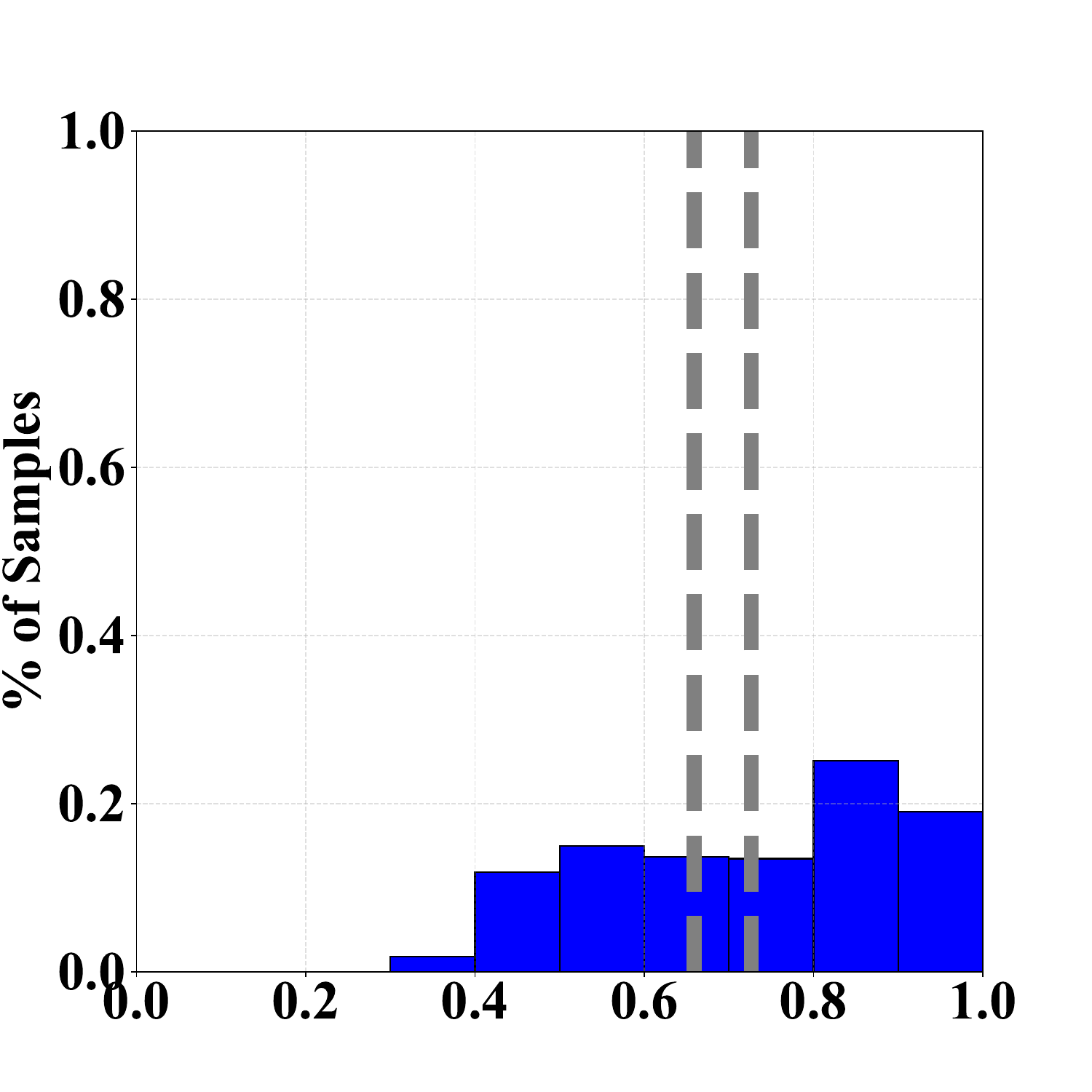}
  }
  \subfigure[fNIRSNet on UFFT Dataset]{
  \includegraphics[width=1.55in]{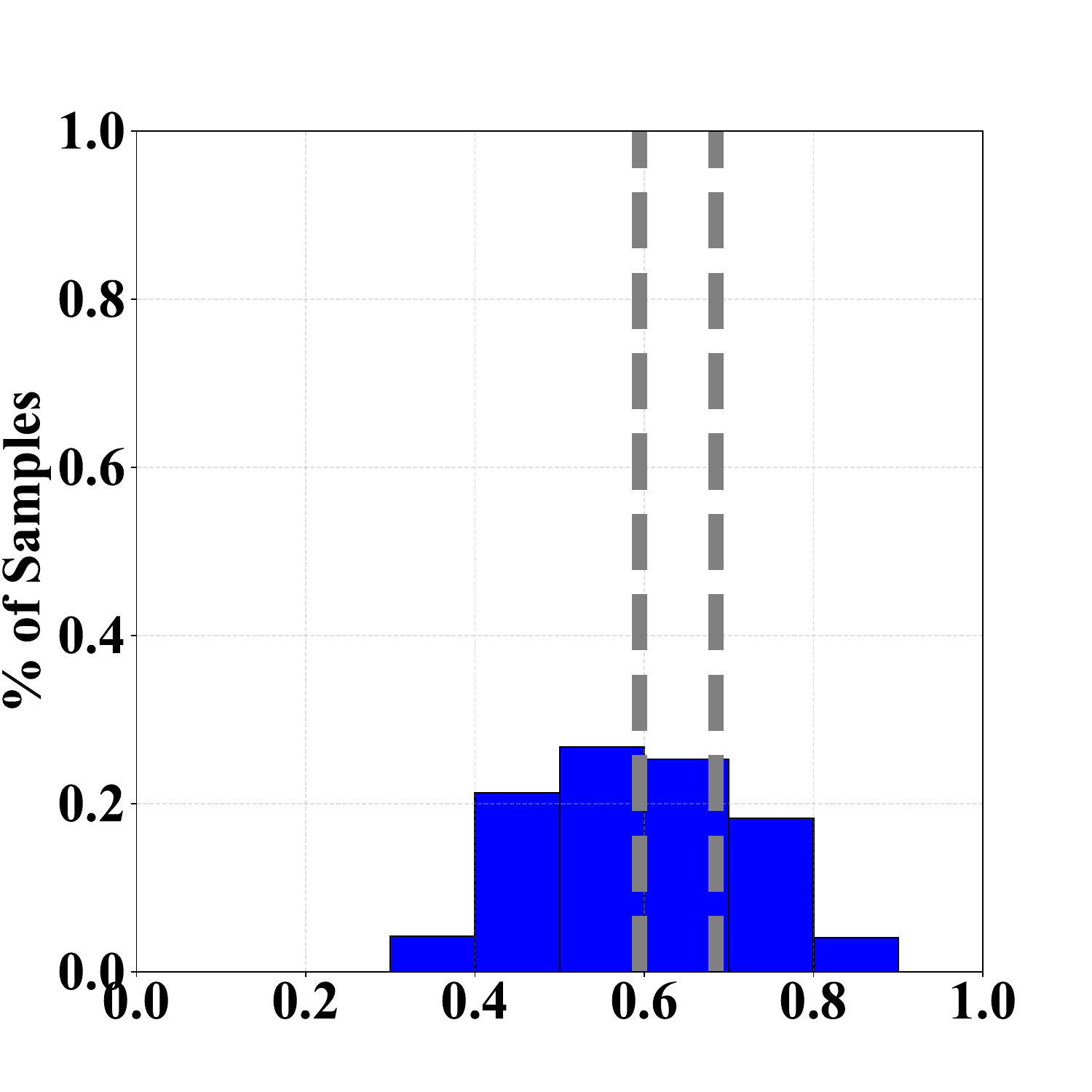}
  }
  \vspace{-0.1cm}
  \caption{CNN+LSTM and fNIRSNet on UFFT Dataset}
  \label{fig:UFFT}
  \vspace{-0.5cm}
\end{figure}

\section{PRACTICAL SKILLS\label{sec:skills}}

\subsection{Balancing Accuracy and Calibration} 

Obtaining a model with both good calibration performance and high accuracy simultaneously can be challenging. Although we cannot establish causality, based on experiments, if we aim for slightly improved accuracy along with better calibration performance in complex tasks, we can employ SCE, ACE, and TACE as evaluation metrics. Additionally, when seeking to assess both calibration performance and accuracy, one effective approach is to consider its AvUC value \cite{krishnan2020improving}. However, due to its complexity, we propose a simpler yet user-friendly method to strike a balance between accuracy and calibration performance.

\begin{equation}
\label{sce_equation}
Score = (1-\alpha) \frac{ACC}{ACC_{baseline}} + \alpha e^{CAL_{baseline}-CAL},
\end{equation}
where $ACC$ represents the mean accuracy, $ACC_{baseline}$ represents the expected accuracy baseline, $CAL$ represents the calibration performance (such as ECE, ACE, etc.), and $CAL_{baseline}$ represents the expected calibration performance baseline. The parameter $\alpha$ is used to measure the emphasis placed on accuracy versus calibration performance.

For example, with $\alpha=0.6$, we assign $ACC_{baseline}$ the maximum value and $CAL_{baseline}$ the minimum value from them. The results are presented in Table \ref{tab:balance_ma} and \ref{tab:balance_ufft}, indicating the superior performance of fNIRSNet.

\begin{table}[hptb]
\caption{BALANCE SCORE OF MA DATASET}
\label{tab:balance_ma}
\vspace{-0.5cm}
\begin{center}
\begin{tabular}{c|c|c|c|c}
\hline
  & \makecell{Score $\uparrow$ \\ @ ECE} & \makecell{Score $\uparrow$ \\ @ MCE} & \textcolor{red}{\makecell{Score $\uparrow$ \\ @ SCE}} & \textcolor{red}{\makecell{Score $\uparrow$ \\ @ TACE}} \\
\hline
LSTM  & 0.85 & 0.80 & 0.92 & 0.92 \\
\hline
fNIRSNet  & \textbf{1} & \textbf{1} & \textbf{0.98} & \textbf{0.98} \\
\hline
\end{tabular}
\end{center}
\end{table}

\begin{table}[hptb]
\vspace{-0.4cm}
\caption{BALANCE SCORE OF UFFT DATASET}
\label{tab:balance_ufft}
\vspace{-0.3cm}
\begin{center}
\begin{tabular}{c|c|c|c|c}
\hline
  & \makecell{Score $\uparrow$ \\ @ ECE} & \makecell{Score $\uparrow$ \\ @ MCE} & \textcolor{red}{\makecell{Score $\uparrow$ \\ @ SCE}} & \textcolor{red}{\makecell{Score $\uparrow$ \\ @ TACE}} \\
\hline
CNN+LSTM  & 0.98 & 0.98 & 0.97 & 0.98 \\
\hline
fNIRSNet  & \textbf{0.99} & \textbf{0.99} & \textbf{1} & \textbf{1} \\
\hline
\end{tabular}
\end{center}
\vspace{-0.4cm}
\end{table}

\subsection{Model Capacity Selection} 

Although very deep or wide models tend to generalize better than smaller models and can easily fit the training set, it's evident that these increases may negatively impact the model's calibration. For instance, the model capacity of CNN+LSTM exceeds that of other models. Despite CNN+LSTM's satisfactory calibration performance in the more complex task of three classifications on UFFT dataset, fRNISNet still exceeds it significantly in terms of accuracy and calibration. Moreover, in the simpler task of two classifications on MA dataset, the superior generalization and accuracy of CNN+LSTM do not offer substantial benefits and also affect calibration. As indicated in Table \ref{tab:ma_result}, smaller models like LSTM (10.92k) and fRNISNet (0.50k) \cite{wang2023rethinking} demonstrate greater benefits in both calibration and classification in the fNIRS field. Therefore, we recommend the utilization of smaller models to ensure certain calibration performance in the fNIRS domain.

\subsection{Temperature Scaling} 

\begin{figure}[htpb]
  \centering
  \vspace{-0.6cm}
  \subfigure[fNIRSNet (Origin)]{
  \includegraphics[width=1.55in]{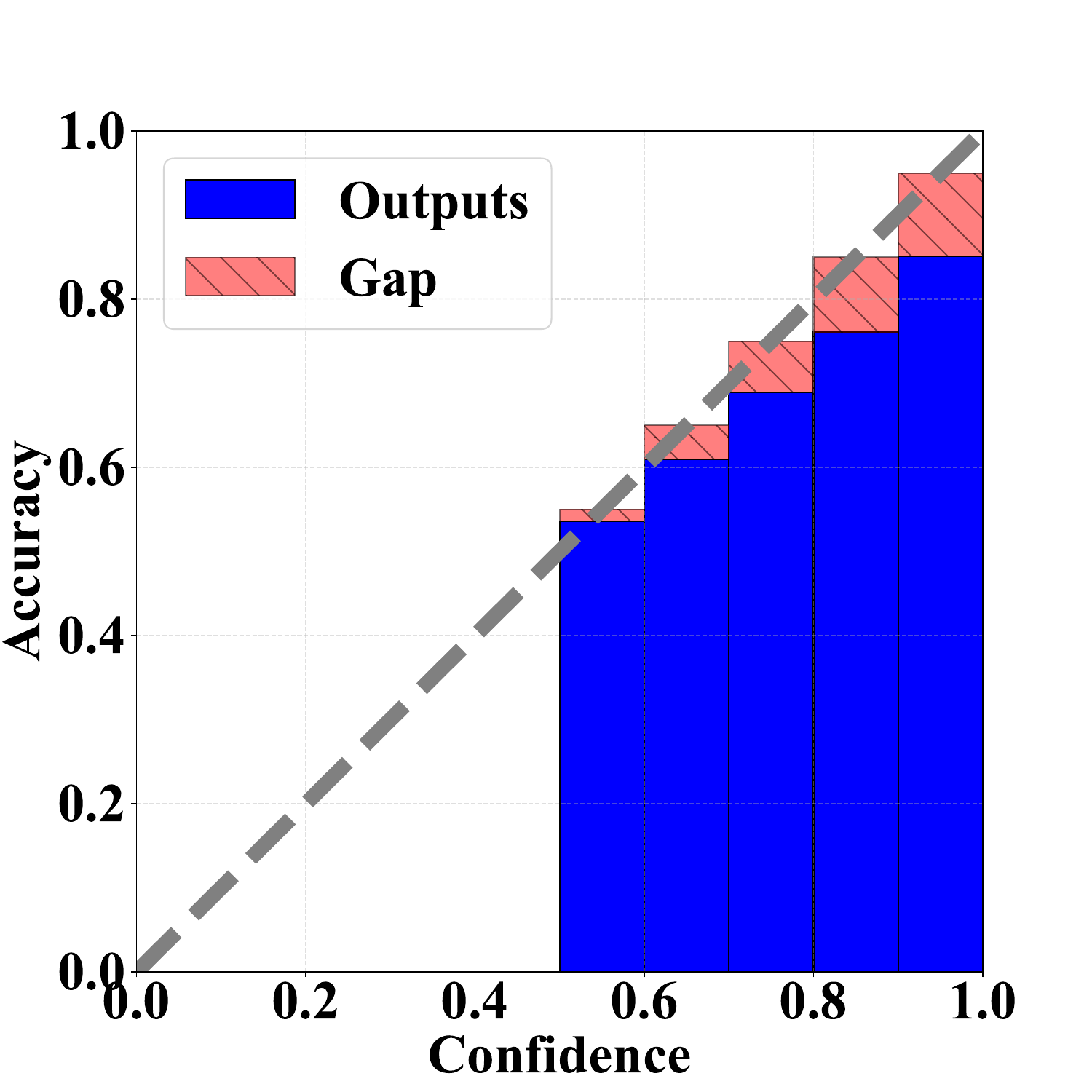}
  }
  \subfigure[fNIRSNet (Temperature Scaling)]{
  \includegraphics[width=1.55in]{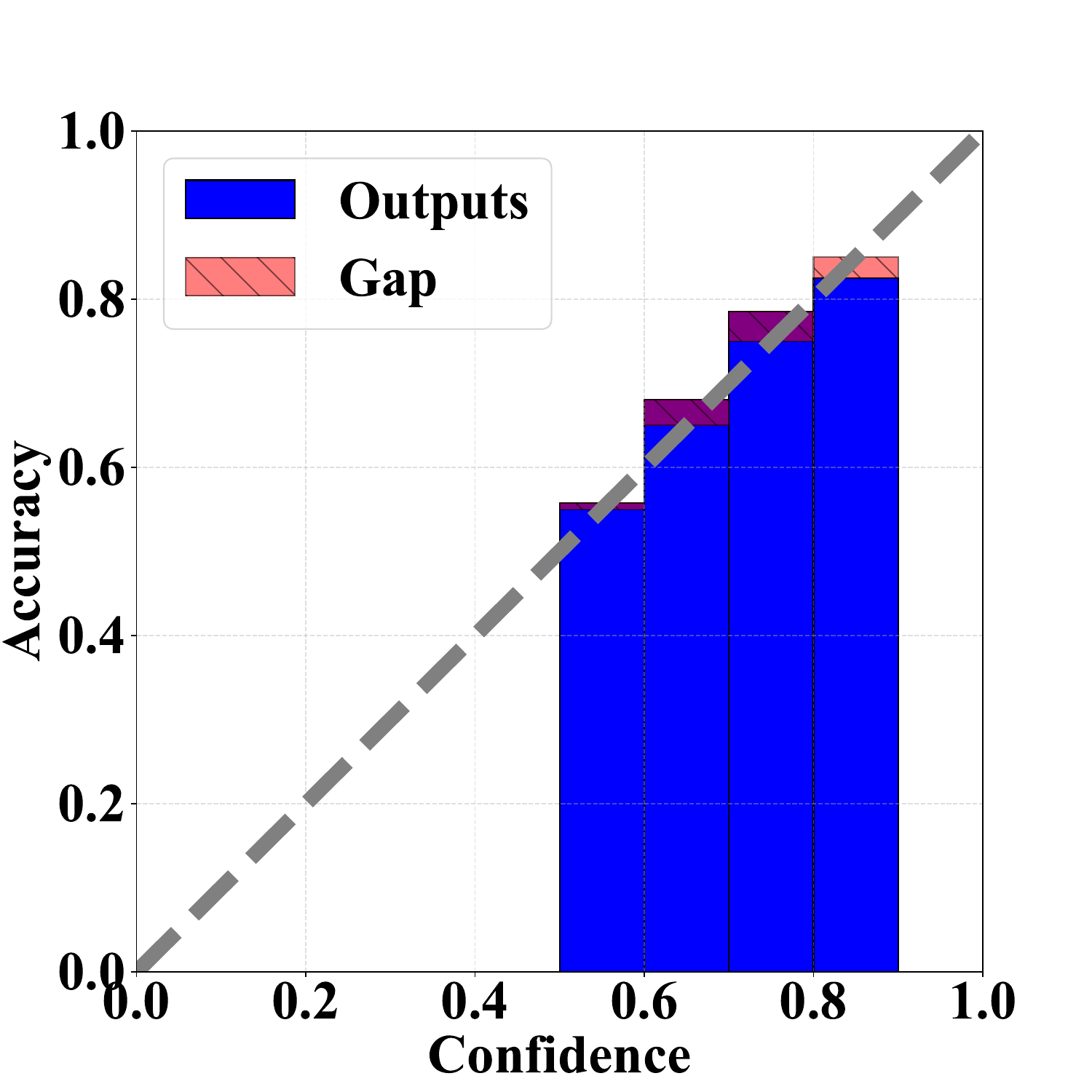}
  }
  \vspace{-0.1cm}
  \caption{Temperature Scaling Experiment on MA Dataset}
  \label{fig:temperature_scaling}
  \vspace{-0.52cm}
\end{figure}

Temperature scaling is an excellent post-porcessing technique for reducing the calibration error of the model, which is the simplest extension of Platt scaling. It uses a single scalar parameter $T > 1$ for all classes. Given the logit output vector $z_i$, the new confidence prediction is as follow.

\begin{equation}
\label{ace_equation}
\hat{q_i} = \max_k \sigma_{SM} (z_i/T)^{k},
\end{equation}
where $T$ is called the temperature, and it softens the softmax with $T > 1$. According to \cite{guo2017calibration} the parameter $T$ does not change the maximum of the softmax function, temperature scaling does not affect the model’s accuracy. However, in our experiments, we discover that although there is a decrease in accuracy, it remains within an acceptable range, and the calibration error is reduced as expected. We demonstrate that the calibration performance of fNIRSNet after being subjected to temperature scaling, which is as shown in Figure \ref{fig:temperature_scaling} and Table \ref{tab:temperature_scaling}.

\begin{table}[hptb]
\caption{TEMPERATURE SCALING EXPERIMENT ON MA DATASET}
\label{tab:temperature_scaling}
\vspace{-0.3cm}
\begin{center}
\begin{tabular}{c|c|c|c|c|c}
\hline
  & ECE & MCE & OE & \textcolor{red}{\textbf{SCE}} & \textcolor{red}{\textbf{TACE}} \\
\hline
\makecell{fNIRSNet \\ (Origin)}  & 0.07 & 0.09 & 0.06 & 0.29 & 0.29 \\
\hline
\makecell{fNIRSNet \\ (Temperature Scaling)}  & \textbf{0.02} & \textbf{0.03} & \textbf{0.00} & \textbf{0.21} & \textbf{0.21} \\
\hline
\end{tabular}
\end{center}
\vspace{-0.4cm}
\end{table}

\section{CONCLUSION \label{sec:conclusion}}

In this letter, we propose integrating calibration into the fNIRS field and assessing the reliability of existing deep learning models, addressing issues result from neglecting calibration. We then establish a benchmark for evaluating deep learning reliability in fNIRS classification, and extensively verify calibration errors in current models through experiments. Additionally, we summarize and experimentally confirm three practical tips for well-calibrated performance. Through this letter, we aim to emphasize the critical role of calibration in fNIRS research and argue for enhancing the reliability of deep learning-based predictions in fNIRS classification tasks. In the future, we aim to investigate additional post-processing techniques to improve the robustness of current models and devise neural networks with enhanced reliability.

\addtolength{\textheight}{-12cm}

\bibliography{references}
\bibliographystyle{unsrt}

\begin{table*}[htpb]
\caption{RESULT OF MA DATASET}
\label{tab:ma_result}
\begin{center}
\begin{tabular}{c|c|c|c|c|c|c|c|c|c|c|c|l}
\hline
 & $\uparrow$  & $\uparrow$ & $\uparrow$ & $\uparrow$ & $\uparrow$ & $\downarrow$ & $\downarrow$ & $\downarrow$ & $\downarrow$ & $\downarrow$ & $\downarrow$ & \\
\hline
Model & Acc.\%  & Precision\% & Rec.\% & F1-score\% & Kappa & ECE & MCE & OE & \textcolor{red}{\textbf{SCE}} & \textcolor{red}{\textbf{ACE}} & \textcolor{red}{\textbf{TACE}} & Reliability Diagrams \\
\hline
\makecell{1D-CNN \\ \cite{sun2020novel}} &
\makecell{62.6 \\ ± 13.9} &
\makecell{64.1 \\ ± 15.8} &
\makecell{60.2 \\ ± 18.1} &
\makecell{61.0 \\ ± 15.5} &
\makecell{0.25 \\ ± 0.28} &
0.28 & 0.33 & 0.25 & 0.41 & 0.40 & 0.40 &
\begin{minipage}[b]{0.15\columnwidth} \centering \raisebox{-.5\height} {\includegraphics[width=80px]{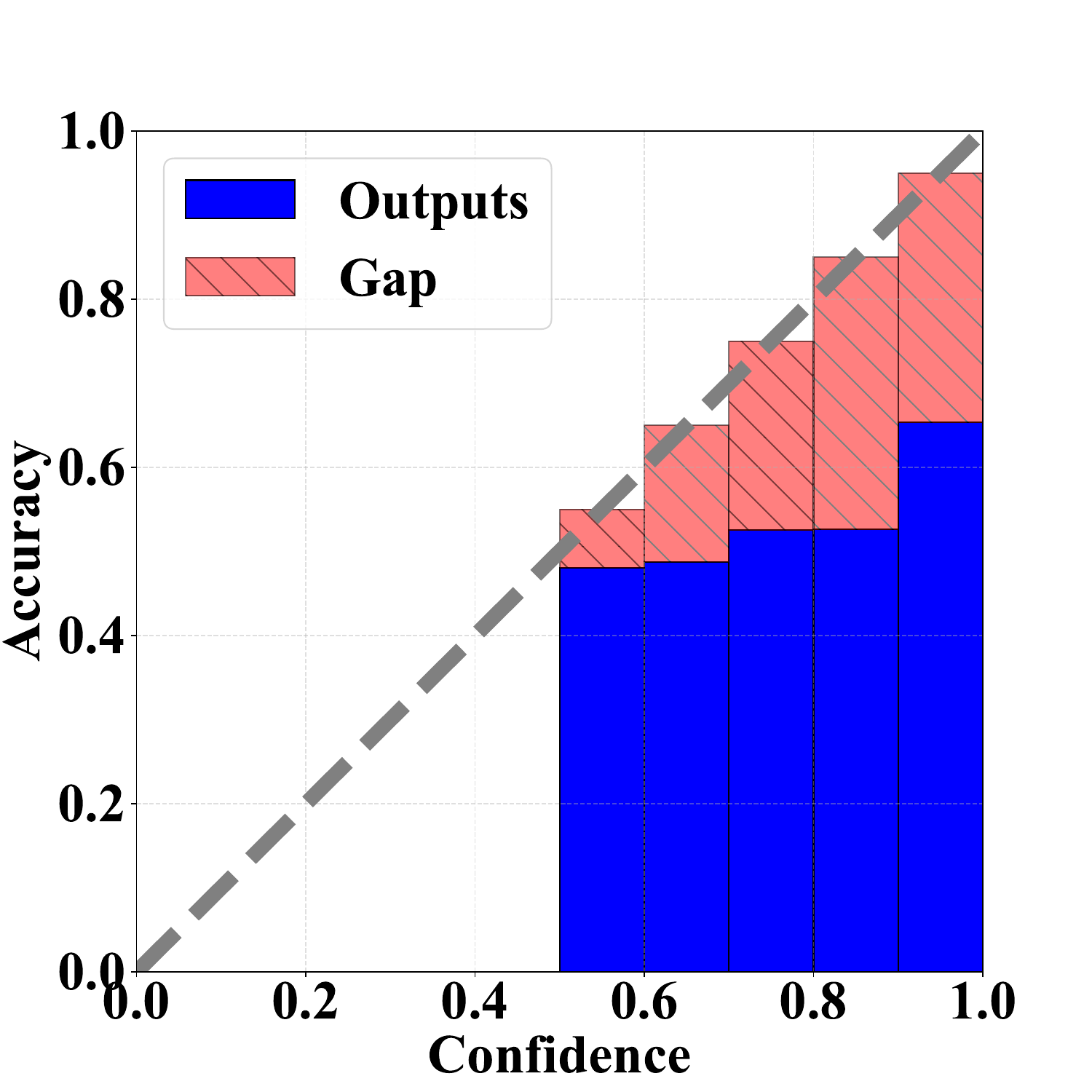}} \end{minipage}\\

\makecell{CNN \\ \cite{lyu2021domain}} & 
\makecell{64.4 \\± 13.9} &
\makecell{65.8 \\± 15.5} &
\makecell{61.7 \\± 17.8} &
\makecell{62.9 \\± 15.2} &
\makecell{0.25 \\ ± 0.28} &
0.17 & 0.24 & 0.15 & 0.32 & 0.32 & 0.31 &
\begin{minipage}[b]{0.15\columnwidth} \centering \raisebox{-.5\height} {\includegraphics[width=80px]{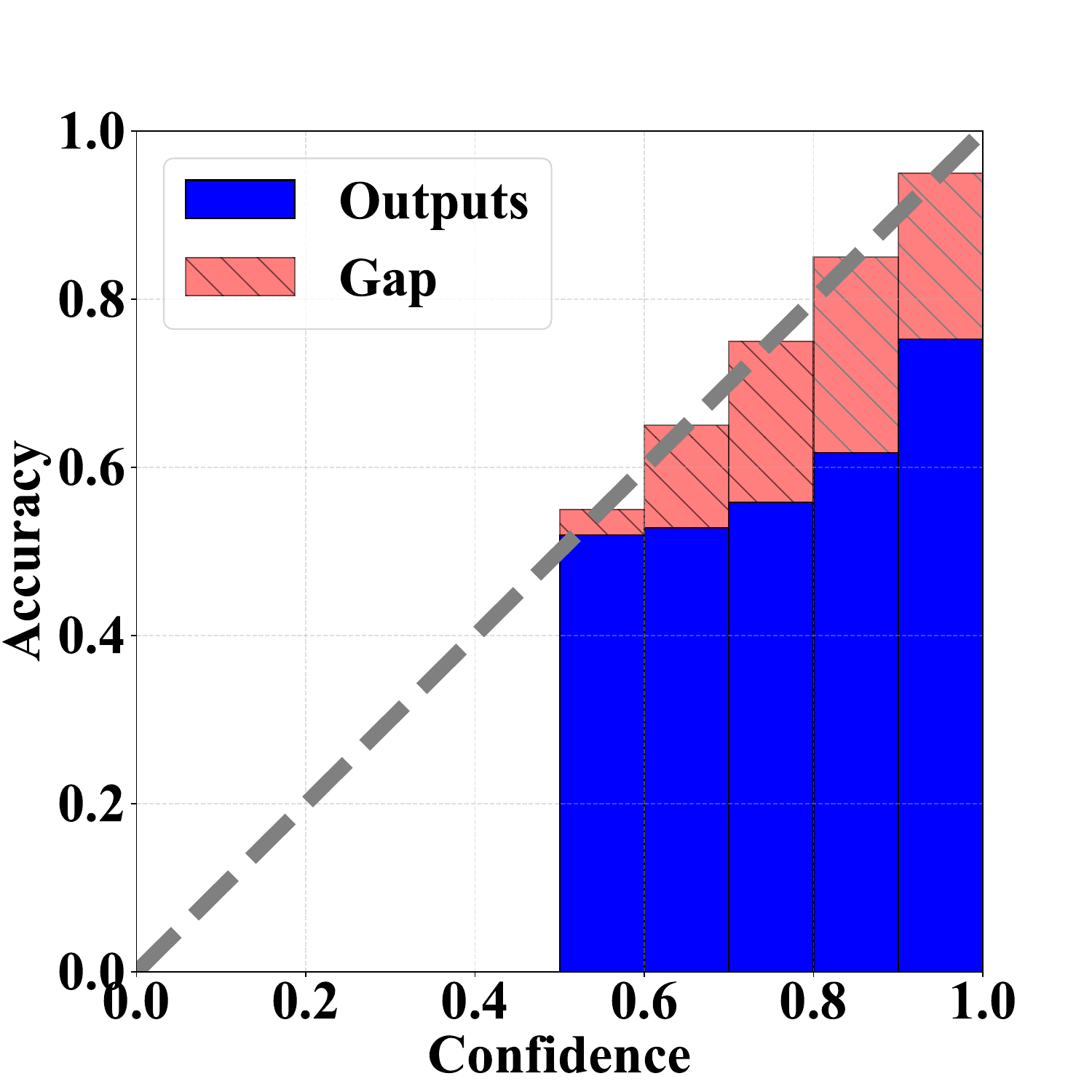}} \end{minipage}\\

\makecell{LSTM \\ \cite{lyu2021domain}} & 
\makecell{56.9 \\± 11.7} &
\makecell{57.7 \\± 12.9} &
\makecell{55.4 \\± 17.2} &
\makecell{55.5 \\± 13.7} &
\makecell{0.14 \\± 0.23} &
0.19 & 0.30 & 0.16 & \textbf{0.26} & \textbf{0.26} & \textbf{0.26} &
\begin{minipage}[b]{0.15\columnwidth} \centering \raisebox{-.5\height} {\includegraphics[width=80px]{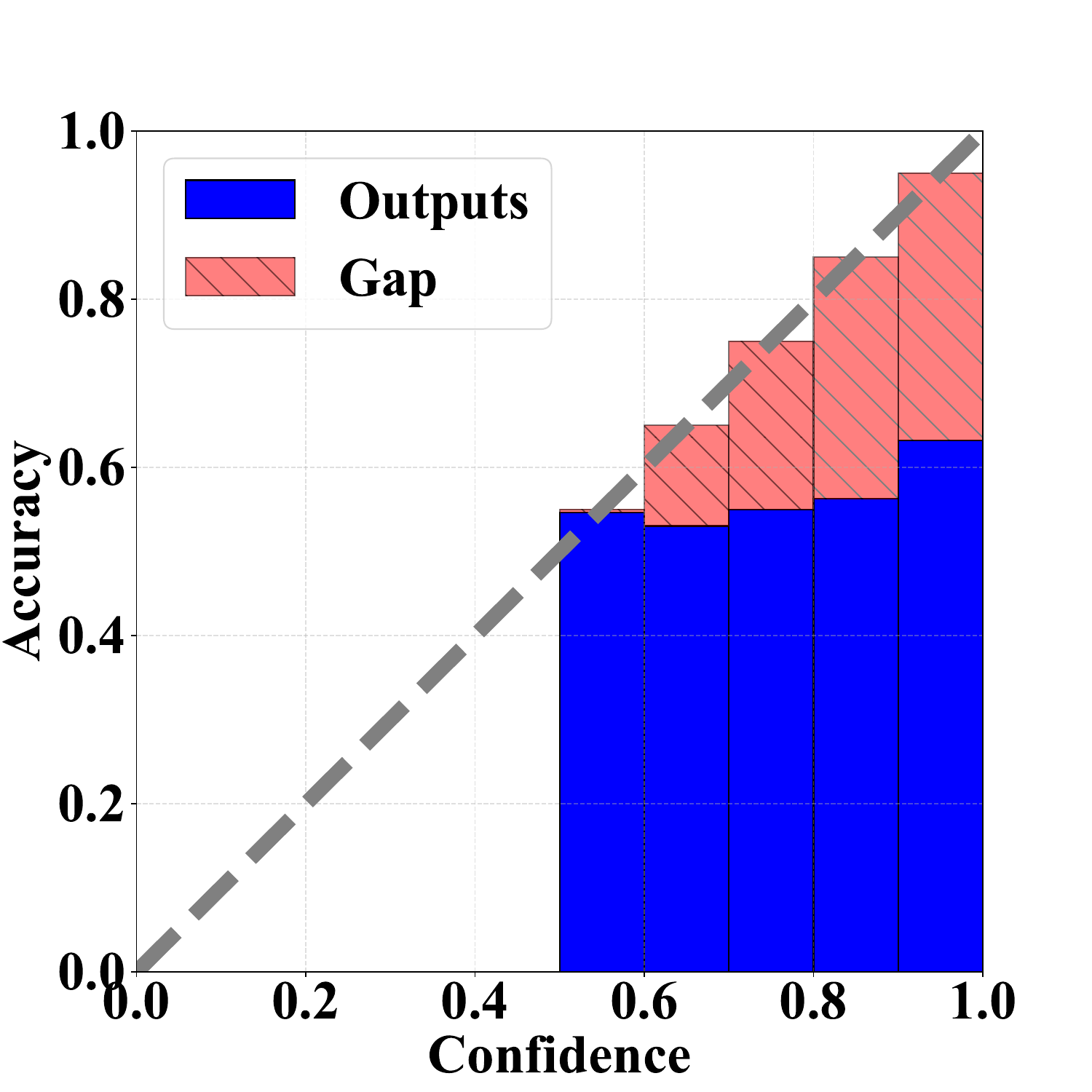}} \end{minipage}\\

\makecell{CNN+LSTM \\ \cite{mughal2021fnirs}} & 
\textbf{\makecell{72.8 \\ ± 12.2}} &
\textbf{\makecell{74.6 \\ ± 13.4}} &
\textbf{\makecell{71.0 \\ ± 16.8}} &
\textbf{\makecell{71.8 \\ ± 13.2}} &
\textbf{\makecell{0.46 \\ ± 0.24}} &
0.16 & 0.36 & 0.10 & 0.37 & 0.34 & 0.34 &
\begin{minipage}[b]{0.15\columnwidth} \centering \raisebox{-.5\height} {\includegraphics[width=80px]{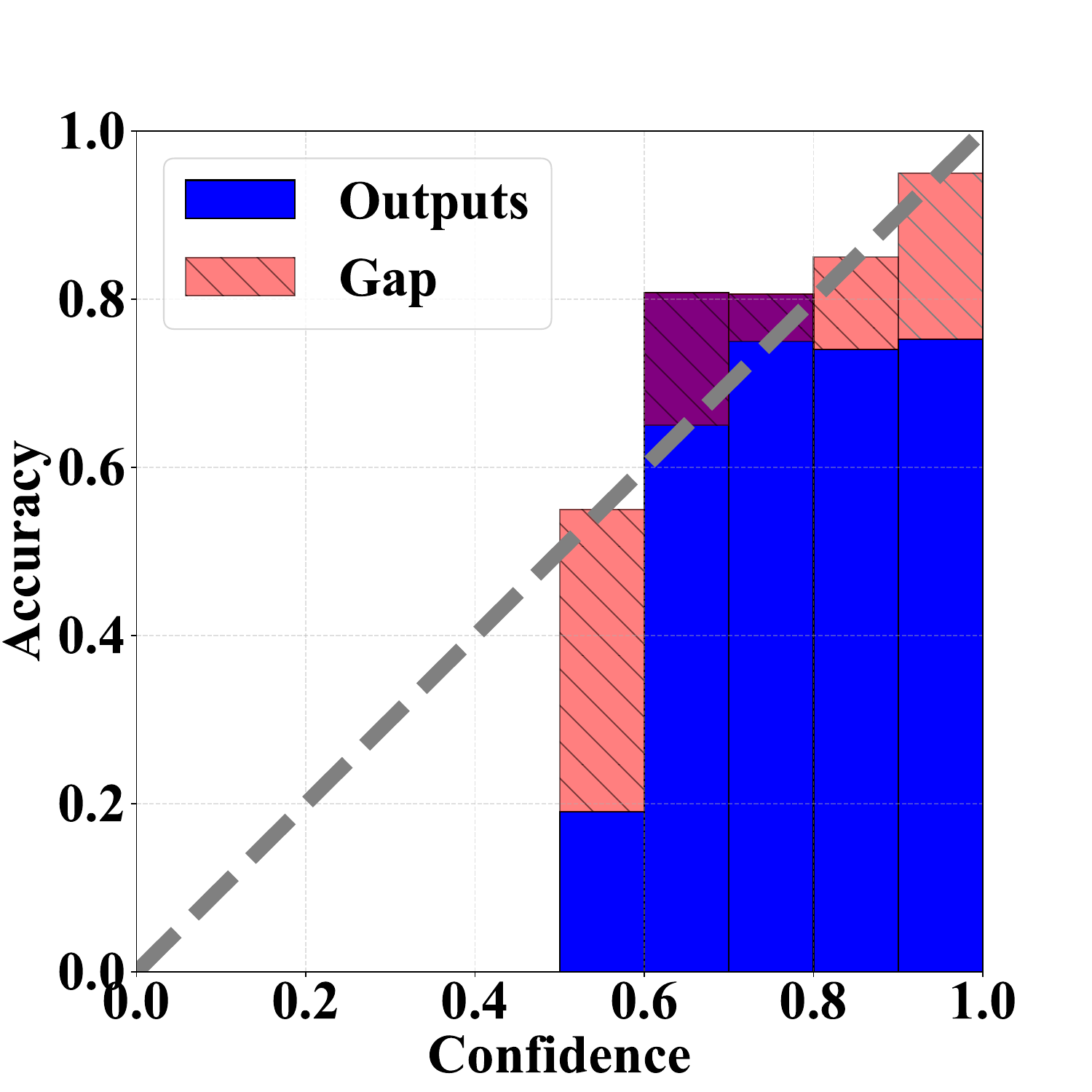}} \end{minipage}\\

\makecell{fNIRS-T \\ \cite{wang2022transformer}} & 
\makecell{64.7 \\± 14.1} &
\makecell{65.7 \\± 16.0} &
\makecell{64.2 \\± 18.5} &
\makecell{63.9 \\± 15.5} &
\makecell{0.29 \\± 0.28} &
0.25 & 0.31 & 0.23 & 0.40 & 0.40 & 0.40 &
\begin{minipage}[b]{0.15\columnwidth} \centering \raisebox{-.5\height} {\includegraphics[width=80px]{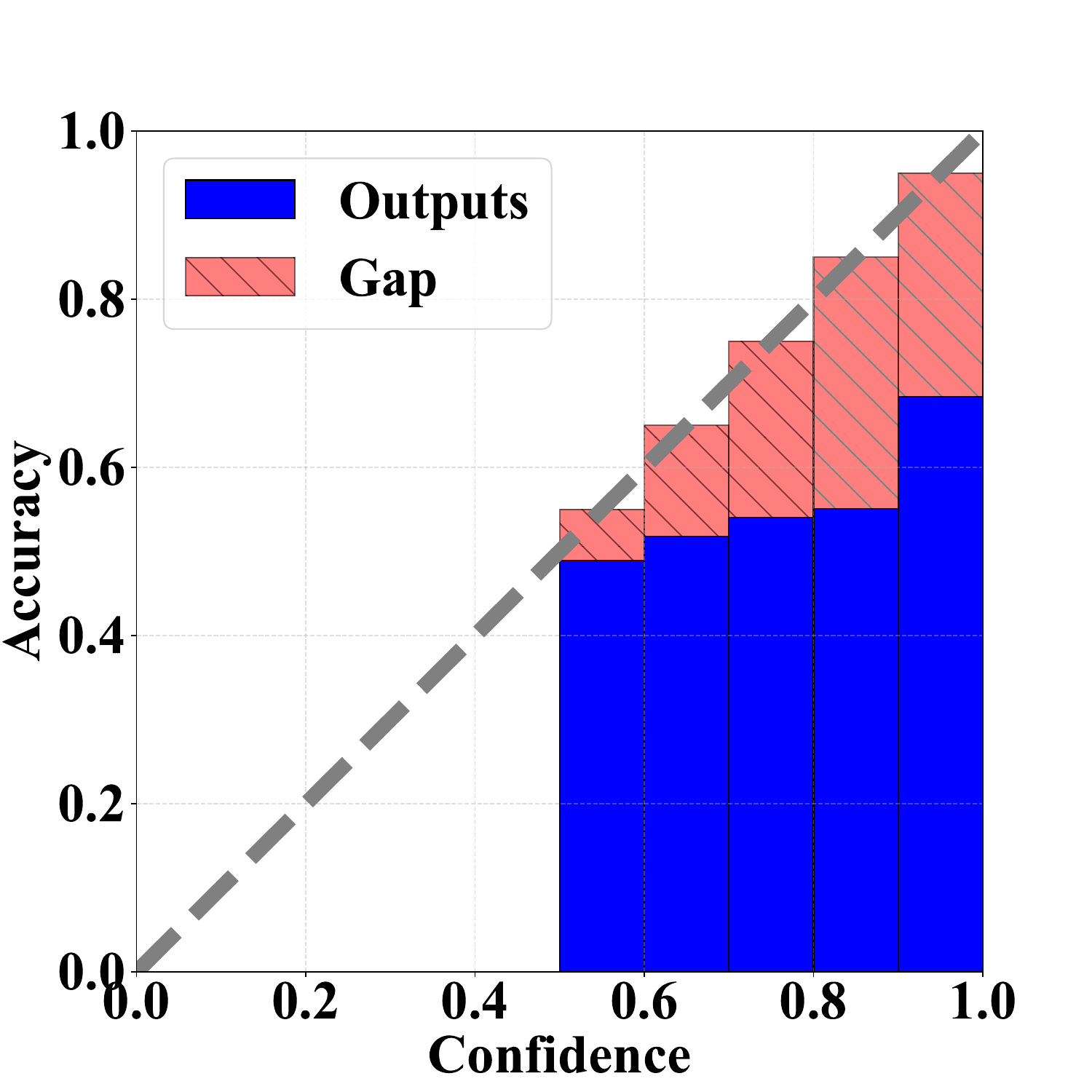}} \end{minipage}\\

\makecell{fNIRSNet \\ \cite{wang2023rethinking}} & 
\makecell{71.9 \\ ± 14.2} &
\makecell{73.5 \\ ± 15.9} &
\makecell{70.1 \\ ± 18.1} &
\makecell{70.9 \\ ± 15.6} &
\makecell{0.44 \\ ± 0.28} &
\textbf{0.07} & \textbf{0.09} & \textbf{0.06} & 0.29 & 0.29 & 0.29 &
\begin{minipage}[b]{0.15\columnwidth} \centering \raisebox{-.5\height} {\includegraphics[width=80px]{Figure/fNIRSNet.pdf}} \end{minipage}\\

\hline
\end{tabular}
\end{center}
\end{table*}

\begin{table*}[htpb]
\caption{RESULT OF UFFT DATASET}
\label{tab:ufft_result}
\begin{center}
\begin{tabular}{c|c|c|c|c|c|c|c|c|c|c|c|l}
\hline
 & $\uparrow$  & $\uparrow$ & $\uparrow$ & $\uparrow$ & $\uparrow$ & $\downarrow$ & $\downarrow$ & $\downarrow$ & $\downarrow$ & $\downarrow$ & $\downarrow$ & \\
\hline
Model & Acc.\% & Precision\% & Rec.\% & F1-score\% & Kappa & ECE & MCE & OE & \textcolor{red}{\textbf{SCE}} & \textcolor{red}{\textbf{ACE}} & \textcolor{red}{\textbf{TACE}} & Reliability Diagrams \\
\hline
\makecell{1D-CNN \\ \cite{sun2020novel}} &
\makecell{57.6 \\± 17.4} &
\makecell{58.5 \\± 17.8} &
\makecell{58.1 \\± 17.3} &
\makecell{56.7 \\± 17.8} &
\makecell{0.37 \\± 0.26} &
0.28 & 0.40 & 0.24 & 0.57 & 0.56 & 0.56 &
\begin{minipage}[b]{0.15\columnwidth} \centering \raisebox{-.5\height} {\includegraphics[width=80px]{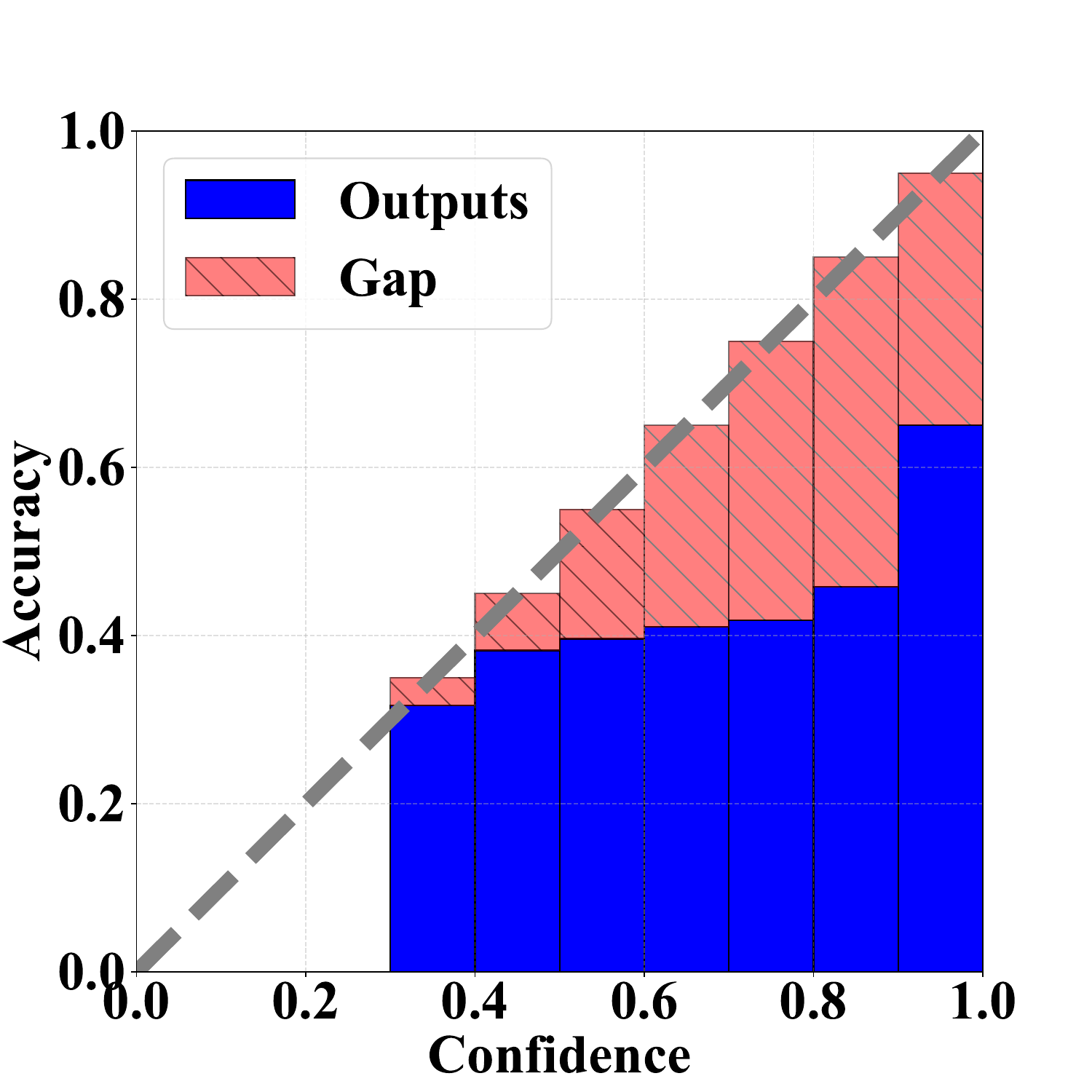}} \end{minipage}\\

\makecell{CNN \\ \cite{lyu2021domain}} & 
\makecell{61.7 \\± 18.6} &
\makecell{62.8 \\± 18.4} &
\makecell{62.0 \\± 18.4} &
\makecell{61.0 \\± 18.7} &
\makecell{0.43 \\ ± 0.28} &
0.15 & 0.20 & 0.12 & 0.51 & 0.51 & 0.50 &
\begin{minipage}[b]{0.15\columnwidth} \centering \raisebox{-.5\height} {\includegraphics[width=80px]{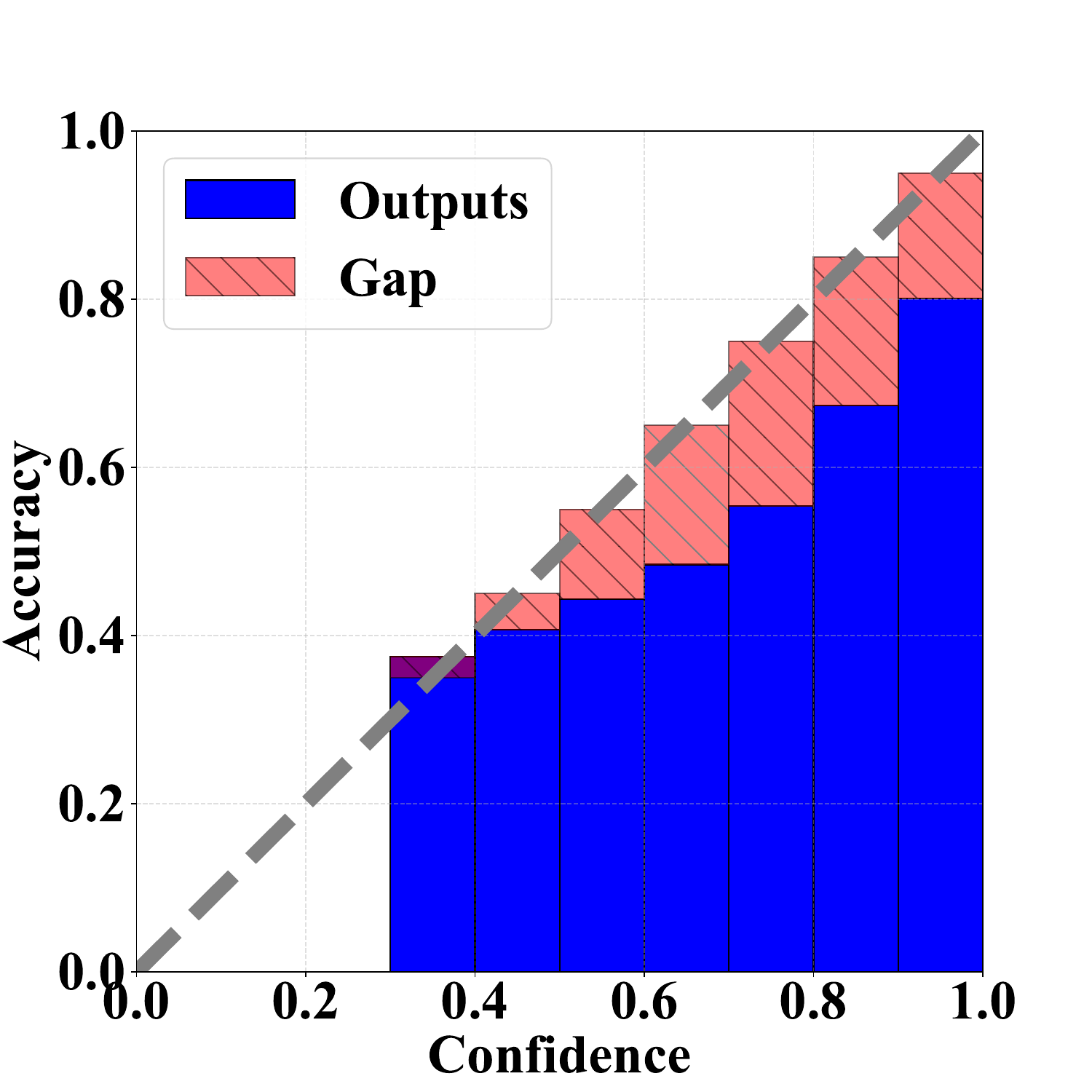}} \end{minipage}\\

\makecell{LSTM \\ \cite{lyu2021domain}} & 
\makecell{46.3 \\± 14.6} &
\makecell{46.8 \\± 14.8} &
\makecell{46.5 \\± 14.7} &
\makecell{45.5 \\± 14.7} &
\makecell{0.20 \\± 0.22} &
0.32 & 0.41 & 0.27 & 0.52 & 0.52 & 0.51 &
\begin{minipage}[b]{0.15\columnwidth} \centering \raisebox{-.5\height} {\includegraphics[width=80px]{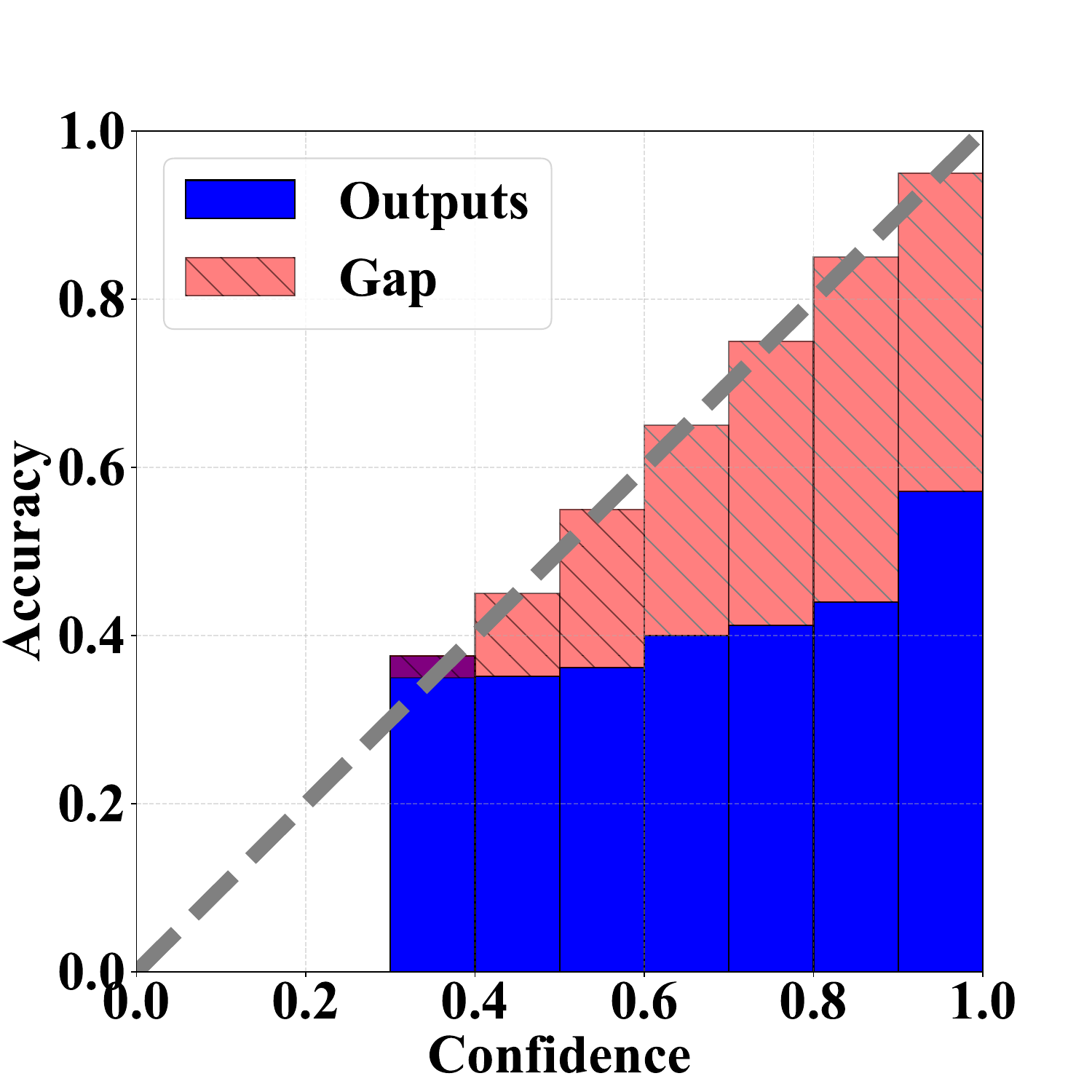}} \end{minipage}\\

\makecell{CNN+LSTM \\ \cite{mughal2021fnirs}} & 
\makecell{65.9 \\ ± 18.0} &
\makecell{67.2 \\ ± 17.9} &
\makecell{66.0 \\ ± 18.0} &
\makecell{65.2 \\ ± 18.3} &
\makecell{0.49 \\ ± 0.27} &
\textbf{0.07} & \textbf{0.11} & 0.05 & 0.49 & 0.48 & 0.47 &
\begin{minipage}[b]{0.15\columnwidth} \centering \raisebox{-.5\height} {\includegraphics[width=80px]{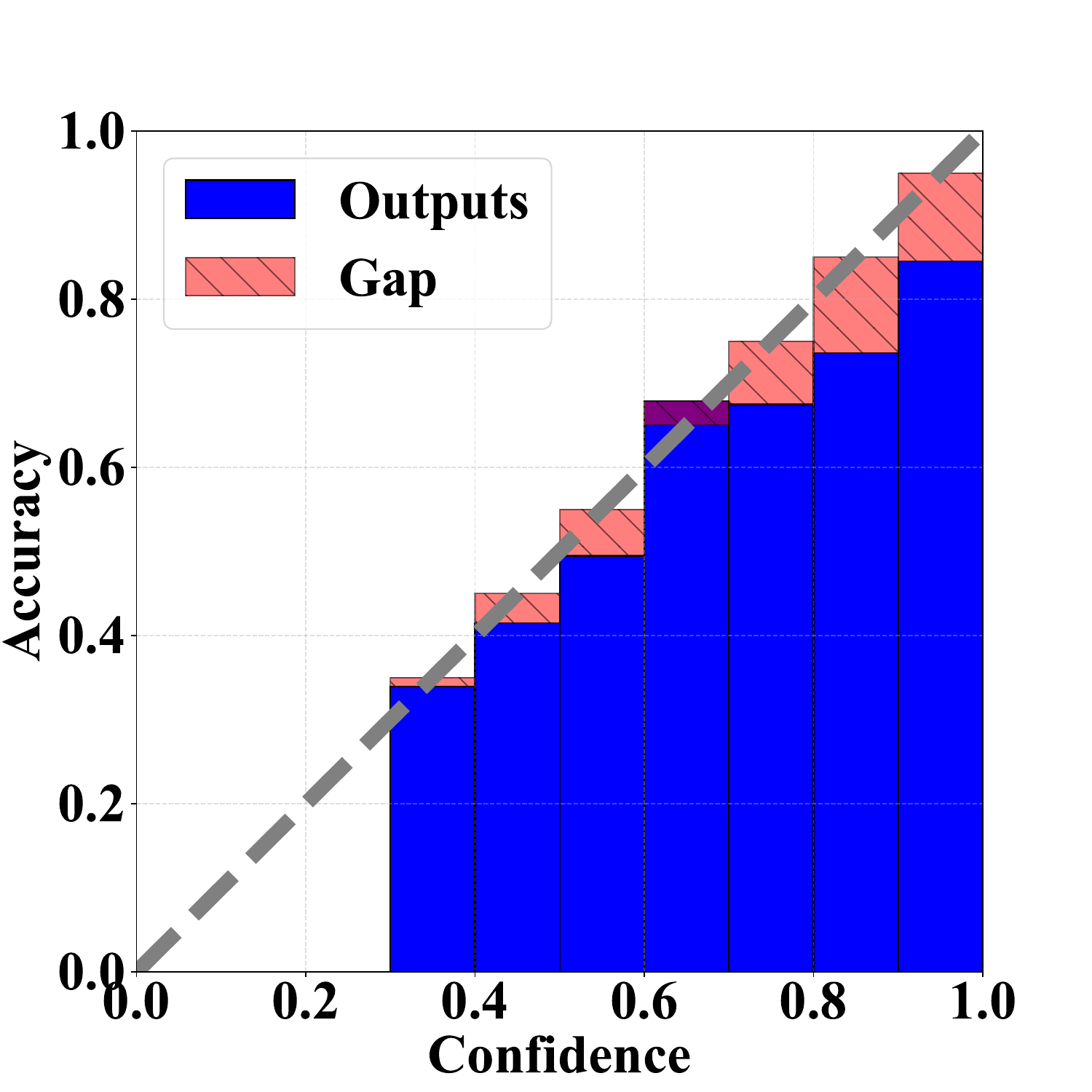}} \end{minipage}\\

\makecell{fNIRS-T \\ \cite{wang2022transformer}} & 
\makecell{61.5 \\± 18.4} &
\makecell{62.4 \\± 18.5} &
\makecell{61.9 \\± 18.1} &
\makecell{60.7 \\± 18.6} &
\makecell{0.42 \\± 0.27} &
0.20 & 0.30 & 0.16 & 0.54 & 0.53 & 0.53 &
\begin{minipage}[b]{0.15\columnwidth} \centering \raisebox{-.5\height} {\includegraphics[width=80px]{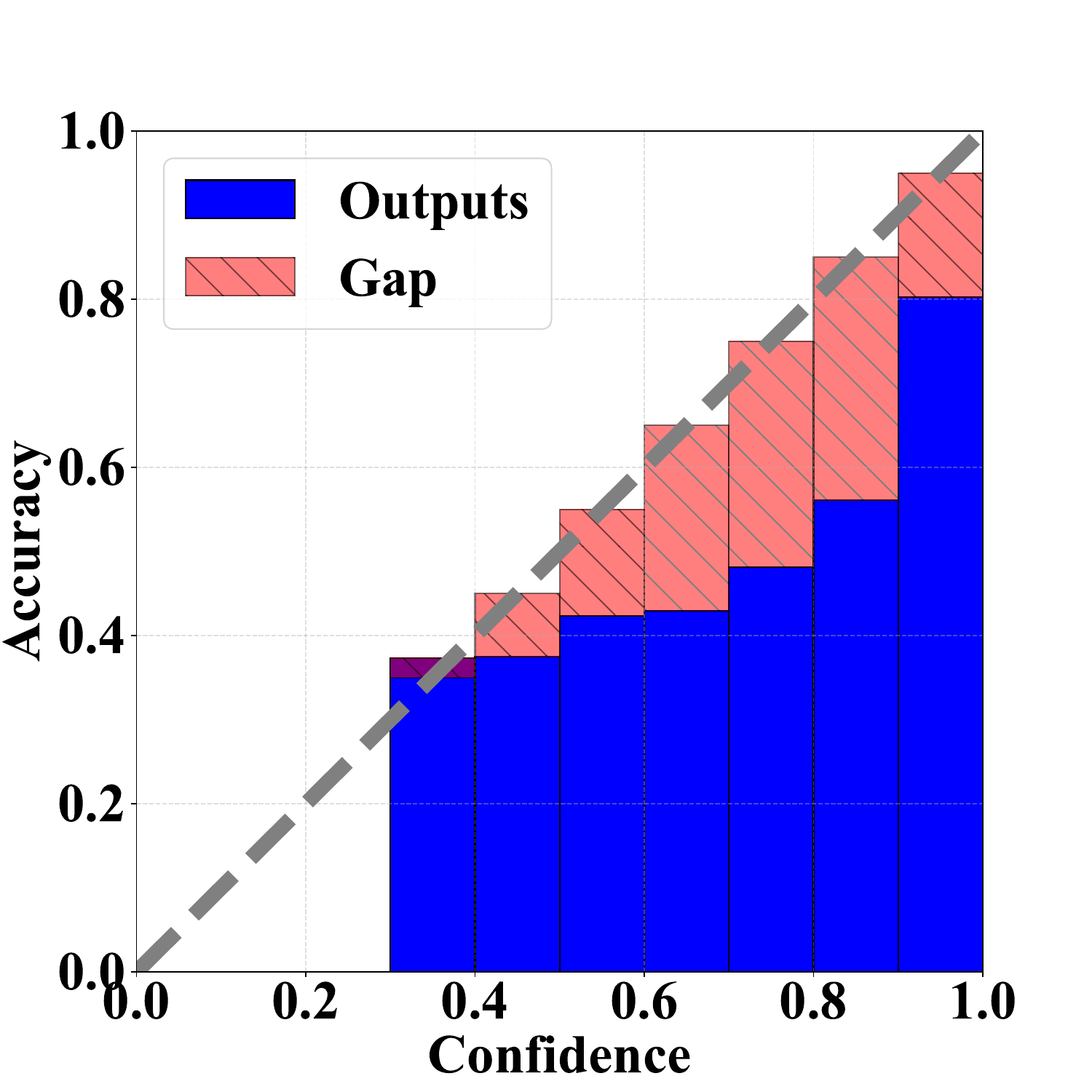}} \end{minipage}\\

\makecell{fNIRSNet \\ \cite{wang2023rethinking}} & 
\textbf{\makecell{68.5 \\ ± 20.1}} &
\textbf{\makecell{69.5 \\ ± 20.1}} &
\textbf{\makecell{68.7 \\ ± 19.9}} &
\textbf{\makecell{67.7 \\ ± 20.4}} &
\textbf{\makecell{0.53 \\ ± 0.30}} &
0.09 & 0.12 & \textbf{0.00} & \textbf{0.46} & \textbf{0.46} & \textbf{0.46} &
\begin{minipage}[b]{0.15\columnwidth} \centering \raisebox{-.5\height} {\includegraphics[width=80px]{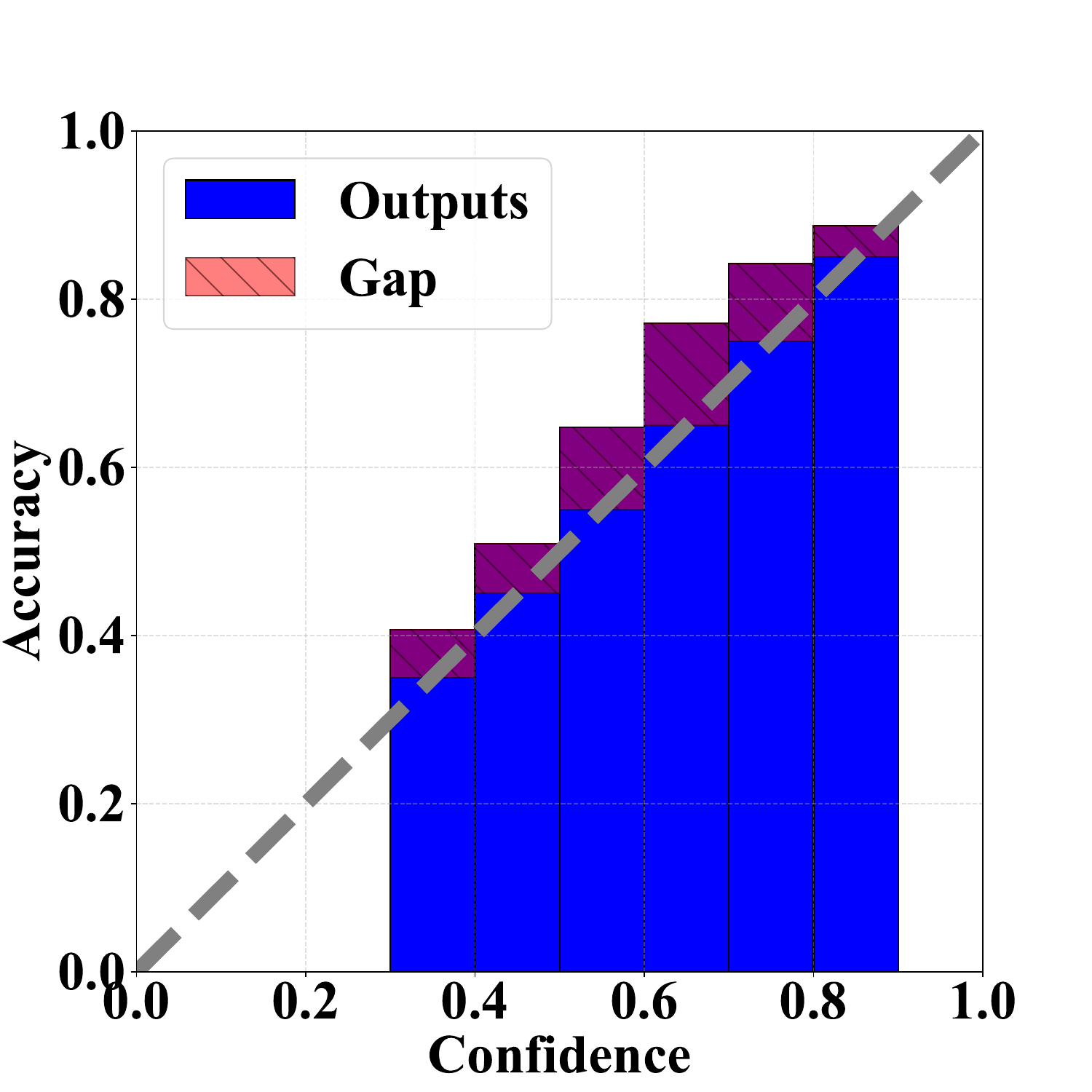}} \end{minipage}\\

\hline
\end{tabular}
\end{center}
\end{table*}

\end{document}